\documentclass[final,3p,times,authoryear]{elsarticle} % nopreprintline
\usepackage[T1]{fontenc}
\usepackage[utf8]{inputenc}
\usepackage{amsmath}
\usepackage{amssymb}
\usepackage{algorithm}
\usepackage{algpseudocode}
\usepackage{booktabs}
\usepackage{tabularx}
\usepackage{graphicx}
\usepackage{microtype}
\usepackage{stfloats}
\usepackage{subcaption}
\usepackage{hyperref}
\usepackage{url}
\hypersetup{colorlinks=true,citecolor=blue,linkcolor=blue,urlcolor=blue}
\graphicspath{{figures/}}
\newcommand{\E}{\mathbb{E}}
\DeclareMathOperator{\KL}{KL}
\DeclareMathOperator{\StopGrad}{StopGrad}
\DeclareMathOperator{\Teacher}{Teacher}
\DeclareMathOperator{\Student}{Student}
\newcommand{\Lpred}{L_{\mathrm{pred}}}
\newcommand{\Ldec}{L_{\mathrm{dec}}}
\newcommand{\Lkd}{L_{\mathrm{KD}}}
\newcommand{\Lppo}{L_{\mathrm{PPO}}}

\journal{Elsevier Journal}

\begin{document}
\begin{frontmatter}
\title{SenseAlign: Learning Decision-Aligned Predictive Representations for Active Urban Sensing with UAVs}

\author[epfl]{Weijiang Xiong}
\ead{weijiang.xiong@epfl.ch}

\author[epfl]{Nikolas Geroliminis\corref{cor1}}
\ead{nikolas.geroliminis@epfl.ch}
\cortext[cor1]{Corresponding author.}

\affiliation[epfl]{organization={Urban Transport Systems Laboratory (LUTS), EPFL},
            city={Lausanne},
            postcode={1015},
            country={Switzerland}}
% Elsevier 5p author-year layout, following GMMJournal/GMM_Traffic.tex.

\begin{abstract}
Active urban sensing orchestrates a mobile sensor fleet to proactively inspect a dynamic road network and gather critical information for system management.
This problem simultaneously requires understanding the complex spatio-temporal evolution and making decisions with partial observations.
In this work, we focus on active sensing in a large-scale urban road network with utilities grounded in dynamic traffic states, while prior work has rarely considered these factors jointly.
We propose \emph{SenseAlign}, a Reinforcement Learning-based (RL) framework that learns decision-aligned predictive representations for coordinating an urban sensing drone fleet.
SenseAlign first trains a privileged teacher agent that exploits future ground-truth traffic for optimal sensing choices.
The student agent has an auxiliary branch that learns to predict future traffic with partial observations, in addition to the RL-trained actor and critic branches.
During student learning, our framework matches the action distributions of the teacher under ground-truth and student-predicted future traffic, while keeping the teacher's observations unchanged.
Such a design shapes the latent space for decision-making instead of prediction alone.
At inference time, the student agent has no overhead as all this learning is done in training.
We conduct comprehensive experiments with the real road network of central Barcelona containing 1570 road segments, and demonstrate that our method consistently outperforms the baselines on multiple utility formulations.
The codebase will be made publicly available upon paper acceptance.
\end{abstract}


\end{frontmatter}
\raggedbottom

\section{Introduction}
\label{sec: introduction}
Urban transportation networks are complex systems with spatio-temporal correlations, and their management builds upon the knowledge of network states and dynamics.
Traditional solutions usually rely on fixed sensor infrastructure, e.g., loop detectors and surveillance cameras, where decent coverage requires substantial investments and continuous maintenance.
Considering the complex spatio-temporal evolution of congestion and the inherent stochasticity of urban traffic system, sparse fixed sensors cannot always provide a reliable monitoring under the varying conditions.
As a result, urban traffic sensing with aerial sensors has been gaining increasing attention in recent years \citep{barmpounakis2020pneuma,fonod2025advanced}.
Drone monitoring can better capture the onset of a traffic breakdown, queue dissipation and propagation of stop-and-go waves. 
They provide a more complete description and more precise measurements of traffic dynamics, particularly for the locations without loop detectors \citep{xiong2025multi}.
These observations are particularly valuable for microscopic traffic modeling, as the detailed trajectories reveal multi-agent interactions such as merging, lane changes, intersection queues, bus stops and turning maneuvers \citep{kim2026lane}.
Those high-quality vehicle trajectory data have supported many studies in microscopic traffic, including parameter extraction \citep{zhou2026itsm, sica2026setting}, trajectory prediction \citep{mohebbi2025vehicle} and traffic flow estimation \citep{song2026high}.
Meanwhile, trajectory data can also enable the research in consistent macro-micro traffic flow modeling \citep{wang2024integrated} and analyzing the capacity drop in the infrastructure \citep{han2026capacity}.
On broader aspects, drone-based observations have created wider social-economic values in emission analysis \citep{pascual2026drone} and sustainability \citep{barmpounakis2025new}.

Although aerial platforms have promising potential in urban sensing, full spatial coverage is hardly possible under realistic resource constraints.
Therefore, the drone fleet must optimize the movements to actively inspect the most valuable areas according to the monitoring requirements.
In this work, we study the problem of active urban sensing, where a swarm of drones moves over a city to monitor network conditions as they evolve.
We consider two types of sensing tasks that are essential for understanding the system dynamics: detecting congestion-change events and estimating the Macroscopic Fundamental Diagram (MFD) \citep{geroliminis2008existence}.
The former aims to observe roads as congestion emerges or dissipates, while the latter requires measurements that represent the aggregate flow and density of the road network.
Both tasks require the fleet to coordinate its movements based on partial observations and uncertain states, since traffic conditions in unobserved areas remain unknown and previously collected measurements become outdated.
Thus, effective sensing depends on understanding how the network evolves and how this evolution changes the value of visiting different locations.

The problem of modeling traffic evolution has long been a central topic in spatio-temporal forecasting, with graph-based methods learning dependencies across connected roads and successive observations \citep{li2018dcrnn}.
For active sensing, such predictive knowledge can help the fleet anticipate traffic conditions at the locations it will reach after moving.
We therefore introduce traffic prediction as an auxiliary task for a reinforcement learning (RL) agent, teaching the model to understand traffic dynamics from partial observations.
Since the ultimate goal is to allocate the drone fleet, prediction accuracy does not strictly translate to the value in decision-making.
For example, a small error near a congestion transition can change which road should be visited, while a larger error at another location may have little influence.
Therefore, an ideal predictor should provide decision-friendly forecasts, following the principle of Smart Predict-then-Optimize (SPO) \citep{elmachtoub2022smart}.

Based on these observations, we propose \emph{SenseAlign}, an RL-based framework that learns decision-aligned predictive representations for drone-based active urban sensing.
A privileged teacher is first trained with access to future ground-truth (GT) traffic, giving it great advantage on the monitoring task \citep{chen2020cheating}.
We then train a student model to make decisions while having no privileged information.
In addition to the standard RL training with Proximal Policy Optimization (PPO) \citep{schulman2017ppo}, the student is augmented with three tasks for representation learning:
\begin{itemize}
\item Future traffic prediction, for understanding the dynamics and propagation of network traffic
\item Knowledge distillation, for imitating the behavior of a privileged model while having only partial information
\item SPO-inspired decision alignment, for preserving the traffic information most relevant to choosing the observations
\end{itemize}
These objectives jointly shape the shared latent space, and allow the learned model to make better monitoring decisions.
Experiments on a large-scale urban road network show that SenseAlign outperforms PPO and the other evaluated baselines on congestion-change monitoring and MFD estimation.
Ablation studies have further verified the contribution of each module, showing the necessity of learning decision-aligned representations.

\section{Related work}
\label{sec: related work}

\subsection{Drone-based monitoring}
\label{subsec: drone-based monitoring}
Drones are favorably utilized in inspection and monitoring across wide application domains.
For landslide inspection, routing models maximize collected geological information while requiring visits to high-priority sites \citep{fang2023landslides}.
\citet{shen2020shipemissions} adapt drone routes to changing vessel locations for air pollution surveillance, and~\citet{liu2026seamonitoring} schedule flights to continuously cover a sea area.
Other applications include infrastructure inspection \citep{yang2025det}, agriculture \citep{tang2025orchard, velusamy2021unmanned}, maritime monitoring \citep{he2026maritime} and wildfire monitoring \citep{merino2012fire}.
Specifically in the transportation domain, drone swarms have been adapted in smart cities for traffic monitoring \citep{khan2020smart}.
With the support of computer vision \citep{fonod2025advanced}, that footage is particularly valuable for analysis of emissions \citep{espadaler2023traffic}, travel behavior \citep{zhu2024real} and road safety \citep{outay2020applications}.

With limited flight resources, drone-based traffic monitoring requires decisions about both spatial coverage and observation value.
\citet{christodoulou2020tour} plan tours that cover a prescribed road network with as few drones as possible under flight-time constraints.
For traffic-responsive navigation, \citet{huang2021decentralized} direct drones to gather around detected road blockages, while \citet{yahia2022estimation} guide a drone toward observations that reduce uncertainty in traffic and incident estimates on a freeway.
For network-level monitoring, \citet{wang2022rushhours} optimize peak-hour monitoring routes using time-varying road profits, \citet{theocharides2024real} aims to estimate per-region MFD with drones, and \citet{chow2016dynamic} formulates repeated visits as information refreshes in a stochastic arc-inventory routing problem.
Learning-based controllers prioritize event-arrival uncertainty under partial observability \citep{khamidehi2021traffic} or combine changing traffic importance with observation age \citep{maljkovic2025monitoring}.
Since drones have become an important factor in urban logistic infrastructure \citep{tan2026uav}, drone monitoring is often considered with other urban applications.
\citet{bai2026TITS} and \citet{xu2023gv} propose to leverage the observations from existing ground vehicles to have better monitoring quality.
\citet{lyu2026multiservice} jointly optimize drone routes for urban monitoring, data collection, and parcel delivery.
\citet{xu2026knowledge} consider both monitoring and parcel delivery drones with rolling-horizon planning, and a follow-up work coordinates local detours to collect road information whose value grows over time \citep{zhu2026detour}.

Despite this progress, existing works either optimize geometric coverage without traffic dynamics, assume the system states are fully known, study traffic-responsive actions on limited road structures, or define monitoring utility according to predefined priorities and known information-growth models.
However, a large-scale urban network can hardly be fully observed, and the predefined scores are not directly connected with the underlying network traffic.
Therefore, we study the monitoring problem under dynamic traffic and partial observability, and define the monitoring utility using the evolving traffic states.

\subsection{Multi-agent planning}
\label{subsec: multi-agent planning}
In general, coordinating a drone fleet to monitor an evolving road network is a multi-agent planning problem, whose central challenge is to define the value of obtained information and plan actions accordingly.

Informative path planning (IPP) aims to plan trajectories for multiple robots to maximize collective information gain in an unknown environment while respecting resource constraints \citep{popovic2024informative}.
Some earlier works have derived their objectives using information theory, such as mutual information, for environment sensing with single or multiple robots \citep{singh2009informative, hitz2014focused}.
Many learning-based methods rely on Gaussian processes (GP) for system modeling and decision-making.
\citet{cao2023catnipp} use a GP-based belief state and train an RL agent to minimize the predicted uncertainty, and \citet{deolasee2024dypnipp} further enhance the framework with a prediction module to anticipate future belief in a dynamic system.
Later, \citet{ma2025dynamic} combine a physics-based model with a linear Gaussian system using Bayesian Filtering to monitor water pollution dispersion.
The objective of IPP is also tightly connected to GP, as many works aim to minimize the uncertainty of the learned GP \citep{zhao2025attention,lew2026aid,yang2023intent}.
Some works instead build an auxiliary reward with GP-based uncertainty reduction, but target discovering a set of static targets in the environment \citep{vashisth2024dynamic,vashisth2026scalable}.

Persistent monitoring plans repeated observations to maintain information about an evolving environment.
Early studies often impose restrictive assumptions that limit the agents' motion space.
For example, \citet{smith2012persistent} optimize the robot motion along predefined loops, and \citet{yu2015persistent} plan cyclic visits, with the visiting order and time intervals subject to optimization.
Their monitoring objectives are often time-dependent accumulation or random event arrivals, without grounding in the actual process of a specific system.
\citet{rezazadeh2021sub} extend the problem scope and consider the distribution of rewards on a spatial graph and optimize paths for multiple mobile agents.
\citet{pinto2022uncertain} assume that the internal states of monitored targets follow linear stochastic dynamics and define observation quality as a function of the distance between agents and targets, thereby optimizing agent trajectories in continuous space.
\citet{zhang2025irada} propose a decentralized optimization method that accounts for drone communication ranges when optimizing their spatial allocation.
In contrast to these optimization-based approaches, \citet{mishra2025galopp} use a graph neural network (GNN) to model inter-agent correlations and train the monitoring agent with multi-agent reinforcement learning (MARL).
Rather than abstract events or information, STAMP and COMPASS focus on monitoring specific moving objects \citep{wang2023stamp,zhang2025compass}.
They combine Gaussian processes with deep learning to model target dynamics and use reinforcement learning to train agents to optimize their sensing positions.

MARL offers a general learning architecture for coordinating agents under distributed observations \citep{yu2022surprising}.
Specifically, MAGPPO combines CNN-based spatial encoders with graph-attention communication to learn persistent coverage policies using penalties that accumulate at unobserved locations \citep{chen2021monitoring}.
The Multi-Agent Transformer (MAT) generates joint actions autoregressively (i.e., one-by-one) instead of simultaneously \citep{wen2022mat}, such that subsequent agents need to consider the actions of previous agents and avoid conflicts.
We adapt MAGPPO and MAT as baselines for the traffic-monitoring tasks.

Compared to existing works, we study the monitoring problem of a partially observable urban traffic system, whose complexity is beyond the capacity of classical models but learnable with deep neural networks \citep{xiong2026unveiling}.
Our monitoring objectives are defined with dynamic traffic states instead of heuristics or static weights, which requires the agent to understand the underlying traffic system rather than exploiting the predefined rules.
Our method takes an RL-based approach and provides insights on how to shape the latent representation space with predictions, and how the predictions can better contribute to decision-making.

\section{Problem formulation and Monitoring Objective}
\label{sec: problem formulation and monitoring objective}

This section will formulate active sensing under the framework of RL (Section~\ref{subsec: reinforcement learning setup}), and define the rewards that are grounded on dynamic system traffic, i.e., changes of road congestion (Section~\ref{subsec: congestion-change event}) and the MFD (Section~\ref{subsec: mfd estimation}).

\subsection{Reinforcement Learning Setup}
\label{subsec: reinforcement learning setup}
In general, we consider an active urban sensing environment where a drone fleet moves over an evolving road network to collect measurements for its specified sensing tasks.
The road network is represented by a graph $G$, where the nodes are road segments and edges represent road connections.
The urban space is discretized into non-overlapping grids, and a drone hovering over a grid cell can obtain perfect traffic information of all roads within.
We aim to coordinate the drones to collect spatially distributed, high-resolution traffic observations at the most informative locations and time.
Throughout the derivation, we use $i$ to index roads, $j$ to index drones, $t$ to index time, and $z$ to index grid cells; bold lowercase and uppercase symbols denote vectors and matrices.

\paragraph{State and Observation} 
Let $\mathbf{X}_t \in \mathbb{R}^{N \times C}$ denote the network-wide traffic state at time $t$, with $N$ road segments and $C$ feature channels per segment.
Its element $\mathbf x_{i,t}$, contains the traffic state of road $i$ at time $t$, where the state has two channels ($C=2$) for traffic flow and density, denoted by $q_{i,t}$ and $k_{i,t}$.
The spatial grid of the road network contains $Z$ non-empty cells in total, and the drone fleet consisting of $M$ drones are not allowed to observe empty cells or overlap on the same cells.
Let $z$ be the index of grid cell, and let $\mathbf c_z\in\mathbb Z^2$ denote the grid coordinates of cell $z$.
Therefore, at each time step, the available new observations are the traffic states of roads covered by the whole drone fleet:
\begin{equation}
O_t=\{\mathbf x_{i,t} : \exists j, \; g(i)=z_{j,t} \},
\label{eq: coverage}
\end{equation}
where $g(i)$ is the index of the grid cell containing road $i$, and $z_{j,t}$ is the cell index occupied by drone $j$ at time $t$.

To provide more context information, the observations $O_t$ are augmented with memories from earlier visits in the same episode and additional metadata.
Concretely, for every road, the complete inputs include the current measured flow and density, time of day, historical traffic, drone positions and coverage mask, last-observed values, observation age, and whether the road has ever been observed (more details in Section~\ref{subsec: model architecture}).
Unobserved traffic states are set to zero before normalization, and they are identified by the binary coverage mask.

\paragraph{Agents and Actions}
The drone fleet relocates synchronously after monitoring current locations for a time window, and the movements follows a centralized policy $\pi(\mathbf{u}_t \mid O_t)$, where $\mathbf{u}_t$ collects the actions from all $M$ drones.
For the next step, a drone $j$ can choose to stay in its current cell $z_{j,t}$ or move to another accessible cell within grid distance $\rho$ by taking action $\mathbf{u}_{j,t} = (\Delta x, \Delta y)$:
\begin{equation}
    \mathbf c_{z_{j,t+1}}=\mathbf c_{z_{j,t}}+\mathbf{u}_{j,t},
    \qquad \sqrt{(\Delta x)^2 + (\Delta y)^2}\leq \rho,
    \label{eq: movement}
\end{equation}
where $\Delta x$ and $\Delta y$ are integers, allowing movement in either direction along each grid axis.
We assume that moving within this range will not affect the monitoring quality of the target cell, and we do not consider the observations during the movements to the target cell, which could be used to improve monitoring quality in future works.
In this work, we focus on the drone allocation problem based on the traffic dynamics, and assume that the deployed drones have sufficient endurance to complete the approximately two-hour monitoring mission.

\paragraph{Reward}
The reward or utility is a numerical description of the monitoring objective, and is also the major signal to guide the improvement of the monitoring agent.
Following the general practice in RL, the objective of active urban sensing is learning a policy $\pi$ to maximize the expected rewards with temporal discount:
\begin{equation}
J(\pi)=\mathbb E_{\pi}\left[\sum_{t=0}^{T-1}\gamma^t r_t\right],
\label{eq: objective}
\end{equation}
where $T$ is the number of decision-making steps and $\gamma$ is a discount factor to prioritize immediate returns.
The reward $r_t$ is given to the action at time $t$, which choses the locations to observe road traffic at the next time step $t+1$.
Contrary to many prior works with hand-crafted rewards, the reward of our active monitoring environment is decided by the live road traffic with complex spatio-temporal dynamics, which introduces more challenges to the monitoring policy. 

\subsection{Congestion-change Event}
\label{subsec: congestion-change event}

Urban traffic congestion typically emerges at certain locations, where the travel demand often exceeds the available capacity, such as intersections, merging areas, or capacity-constrained links.
These bottlenecks can act as initial congestion hotspots, from which congestion progressively propagates across the network through queue spillbacks, upstream shockwaves, and interactions between adjacent links \citep{bellocchi2020unraveling,li2015percolation}, potentially leading to widespread network congestion.
However, this process is a complex spatio-temporal stochastic processes, which can hardly be modeled by deterministic or monotonic spreading.
Urban traffic dynamics are subject to multiple sources of uncertainty, including stochastic fluctuations in travel demand, variations in driving behavior, signal timing and control decisions, route-choice adaptations, and unexpected disruptions such as incidents.
Consequently, individual links may experience repeated transitions between congested and uncongested states as local traffic conditions evolve, especially links that are operating close to capacity.
This dynamic interplay between congestion propagation and temporary recovery creates complex spatiotemporal patterns, in which congestion may intermittently dissipate and re-emerge, even as the overall extent of network congestion continues to grow.

Timely and accurate identification of these congestion-state transitions is essential for effective and proactive traffic management.
Beyond providing a static representation of network conditions, monitoring these transitions enables traffic managers to detect emerging bottlenecks, distinguish persistent congestion from temporary fluctuations, and identify the critical locations that influence network-wide congestion propagation.
Such information supports the early deployment of targeted control measures, including adaptive signal timing \citep{zoabi2026enhanced,zoabi2027stability}, dynamic route guidance \citep{chen2024iterative}, and demand management \citep{balzer2024modal}, before localized disruptions escalate into widespread congestion.
Equally important, detecting transitions from congested to uncongested states allows traffic managers to assess the effectiveness of implemented interventions, identify opportunities for network recovery, and adapt control strategies accordingly.
Ultimately, capturing the evolving spatial and temporal structure of congestion can improve the responsiveness, efficiency, and robustness of urban traffic management, while reducing unnecessary interventions and making more effective use of limited monitoring and control resources.

With those considerations, the congestion-change objective rewards the monitoring agent when the drones observe roads that enter or exit congestion.
Congestion entry and exit labels are constructed offline using a heuristic-based method.
The congestion score for road $i$ at time $t$ is
\begin{equation}
    c_{i,t}=\operatorname{clip}\left(1-{v_{i,t}}/{v_i^{\mathrm{free}}},\,0,\,1\right),
    \label{eq: congestion score}
\end{equation}
where $v_{i,t}$ is the traffic speed and $v_i^{\mathrm{free}}$ is the road's free-flow speed.
All roads are initially uncongested, and at time $t$, an uncongested road enters congestion when the congestion score exceeds 0.5 for the next two consecutive time steps $t+1$ and $t+2$.
Similarly, a congested road exits congestion when its score falls below 0.4 for the next two steps.
The change indicator $d_{i,t}$ equals one when either transition occurs on road $i$ at time $t$, and zero otherwise.
We use a lower threshold for exits, combined with the two-step persistence requirement, to distinguish a brief speed fluctuation from actual congestion release.

Since a state change in busy major roads substantially influences the experience of more road users than minor roads, we introduce a weight factor $w_i$ to prioritize those important roads.
Besides, considering urban congestion can propagate and dissipate dynamically, acquiring knowledge of unobserved areas is more favorable than repeatedly exploiting a few hot spots.
Thus, we apply an information-age factor $f(a_{i,t})$ to encourage obtaining fresh information from under-explored areas.
Concretely, these two factors are defined as follows:
\begin{align}
    w_i&=\bar{q}_i l_i/\bar{l},
    \label{eq: road weight}\\
    f(a)&=p+(1-p)(1-e^{-a/a_0}),
    \label{eq: freshness}
\end{align}
where $\bar{q}_i$ is the average traffic flow of road $i$ in the training data, $l_i$ is the length of road $i$, and $\bar{l}$ is the average road length over the whole network.
The ages are initialized to 20 at episode start, and a road observed at the previous step has zero age.
At the end of each time step, the information age of observed roads are reset to zero and those for unobserved roads will increase by one.
We set $p=0.2$ and $a_0=5$ steps, so staying at the same place will get 20\% normal rewards, and revisiting weight starts at 0.2 from the second next step and approaches one as observations becomes stale.

Finally, the congestion-change reward combines the change indicator $d_i$ with importance factor $w_i$ and information age $f(a_{i,t})$.
The reward for one drone sums up the weighted transitions observed at the next time step $t+1$:
\begin{equation}
    r_{j,t}^{\mathrm{cong}}=\sum_{i\in I_{j,t+1}}w_i d_{i,t+1}f(a_{i,t+1}),
\label{eq: congestion general}
\end{equation}
where $I_{j,t}=\{i:g(i)=z_{j,t}\}$ denotes the roads observed by drone $j$ at time $t$.
The team reward is summed over the individual rewards:
\begin{equation}
r_t^{\mathrm{cong}}=\sum_{j=1}^{M}r_{j,t}^{\mathrm{cong}}
\end{equation}

\subsection{MFD estimation}
\label{subsec: mfd estimation}
Network-level traffic control decisions, such as perimeter control, critically depend on the accurate estimation of the MFD and vehicle density within each region \citep{sirmatel2021modeling,hamedmoghadam2022percolation}.
Since the MFD captures the relationship between regional vehicle density and network production, reliable estimates of the number of vehicles circulating in each region are essential for identifying prevailing traffic conditions, assessing proximity to critical density thresholds, and determining appropriate interregional transfer flows \citep{saeedmanesh2021extended,sirmatel2020nonlinear}.
Errors in density estimation can lead to incorrect assessments of regional congestion levels and, consequently, inappropriate perimeter control actions \citep{chen2024iterative}.
For instance, underestimating density in an already congested region may result in excessive inflows, further aggravating congestion and reducing network throughput, whereas overestimating density may unnecessarily restrict inflows and lead to inefficient utilization of available road capacity, potentially transferring congestion to adjacent regions \citep{lei2026constrained}.
Such errors can propagate across connected regions, undermining network-wide coordination and potentially resulting in traffic conditions worse than those observed without control.
Therefore, accurate and timely estimation of regional road network state is not merely a prerequisite for effective MFD-based perimeter control but a fundamental factor in ensuring the stability, robustness, and overall performance of the traffic management system \citep{leclercq2019macroscopic,gu2022instability}.

Aiming to accurately estimate the road network state, the MFD estimation task rewards the agent according to how well their collective measurements represents network-wide flow and density.
Concretely, at time $t$, the ground-truth road network flow $Q_t$ and density $K_t$ are the length-weighted flow and density of all road segments:
\begin{equation}
    Q_t=\frac{\sum_il_iq_{i,t}}{\sum_il_i},
    \qquad K_t=\frac{\sum_il_ik_{i,t}}{\sum_il_i}.
    \label{eq: true mfd}
\end{equation}
Meanwhile, the estimated $\hat Q_t$ and $\hat K_t$ are averaged over all road segments observed by the drone fleet (denoted by $I_t$):
\begin{equation}
    \hat Q_t=\frac{\sum_{i\in I_t}l_iq_{i,t}}{\sum_{i\in I_t}l_i},
    \quad \hat K_t=\frac{\sum_{i\in I_t}l_ik_{i,t}}{\sum_{i\in I_t}l_i}.
    \label{eq: estimated mfd}
\end{equation}
To jointly consider flow and density estimations, we average their normalized errors as follows:
\begin{equation}
e_t^{\mathrm{MFD}}=\frac{1}{2}\left(\frac{|\hat Q_t-Q_t|}{s_Q}+\frac{|\hat K_t-K_t|}{s_K}\right),
\label{eq: mfd error}
\end{equation}
where $s_Q\approx0.19549$ veh/s and $s_K\approx0.05679$ veh/m are the average flow and density over all training sessions and all road segments.
Then, the MFD estimation reward at time $t$ is the negative estimation error after taking the chosen action, i.e., the error at the next time step $t+1$:
\begin{equation}
r_t^{\mathrm{MFD}}=-e_{t+1}^{\mathrm{MFD}}.
\label{eq: mfd reward}
\end{equation}
While the final congestion-change reward is a sum of individual rewards, the MFD reward is a shared reward computed from the aggregated observations.
The two tasks therefore require complementary forms of coordination: reaching local traffic changes in time and collecting a representative network sample.

\section{Learning method}
\label{sec: learning method}
This section first introduces the overall framework to illustrate the higher-level idea (Section~\ref{subsec: two-stage learning framework}), and then provides further details on model architecture (Section~\ref{subsec: model architecture}) and learning objectives (Section~\ref{subsec: learning objectives}). 

\subsection{Two-stage learning framework}
\label{subsec: two-stage learning framework}

\begin{figure}[!htb]
\centering
\includegraphics[width=\linewidth]{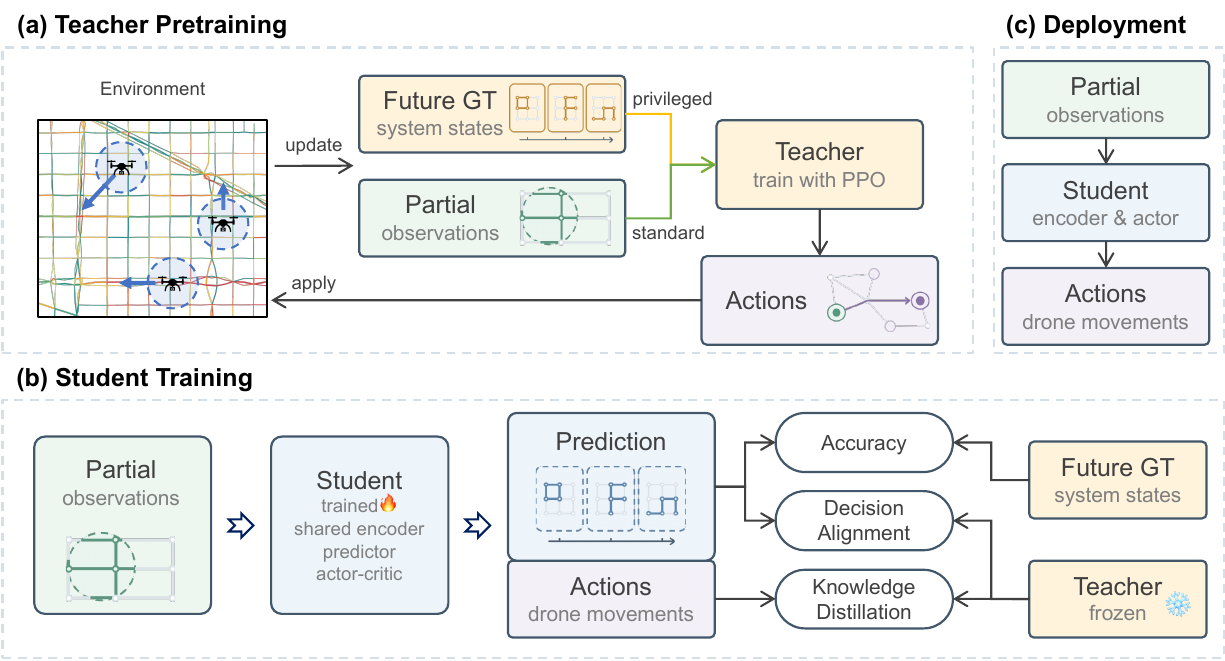}
\caption{Overview of the proposed two-stage training framework and deployment.
(a) A privileged teacher is pretrained with access to ground-truth future traffic.
(b) A student learns to act from partial observations through PPO, where auxiliary supervision from the frozen teacher and future traffic ground truth jointly improves the learned representations.
(c) Deployment uses only the student's encoder and actor, with no additional inference cost from the auxiliary training components.}
\label{fig: system}
\end{figure}

Figure~\ref{fig: system} shows the two-stage learning framework consisting of teacher pretraining and student learning, followed by the deployment.
The teacher first learns to exploit future traffic for better sensing choices, the student then learns with additional supervisions from the teacher and future traffic states while having standard inputs only.
In both learning stages, the data pipeline and training workflow follows the PPO \citep{schulman2017ppo} implemented in Stable-Baselines 3 \citep{stable-baselines3}, and separate policies are trained for the two types of rewards in Section~\ref{sec: problem formulation and monitoring objective}.

\paragraph{Teacher pretraining}
As shown in Figure~\ref{fig: system}(a), the teacher model is an independent PPO agent that knows the future traffic in advance.
At each step, the environment provides the teacher with the standard partial observation (same as the student) and the complete future traffic state (privileged information).
The teacher combines these inputs to select fleet movements, which determine the roads observed at the next step and the resulting monitoring reward.
Since the reward is tightly connected with the future traffic dynamics, knowing the future will give great advantage to the teacher agent.
Because the teacher needs to spend fewer efforts exploring the environment, and can utilize future traffic to efficiently exploit the task knowledge. 
Such privilege can allow the teacher to escape local minima that a student model may struggle with its partial observations, as a student may need to make a trade-off between exploring the environment and exploiting the expected reward.
After pretraining, the teacher model serves as a frozen reference in knowledge distillation and decision alignment during student training, helping the student make better decisions.

\paragraph{Student training}
The student model is an independent PPO-based RL agent trained to make decisions based on the collected partial observations, as shown in Figure~\ref{fig: system}(b).
Since high-quality monitoring depends on properly understanding the complex traffic system, the model has a shared encoder to model the environment, which maps available spatio-temporal observations to a shared feature space for three subsequent branches.
The actor branch selects drone movements, the critic estimates the expected cumulative return from a given state, and the predictor is trained to estimate future system states.
Following the general PPO pipeline, the actor and critic are updated with the obtained rewards, and in addition, the teacher model and future GT provides auxiliary learning signals to improve the learned latent space.
Specifically, the student's predictions are trained to accurately reproduce the future GT, encouraging the student to learn the complex traffic dynamics and preserve traffic-related information in the shared encoder.
Unlike the teacher model who requires future GT as input, the student model does not depend on the future traffic at inference time, and future information is only needed at training time for better representation learning.
Meanwhile, the predictions are also given to the teacher in place of its privileged knowledge (future GT), and the resulting action distribution is matched with the GT-conditioned teacher actions, aligning the predictions with decision values.
Besides, knowledge distillation instructs the student to directly mimic the teacher's behavior, which may guide the student out of its local minima.
Together, these signals teach the student to learn future traffic, preserve decision-critical information, and imitate privileged movements from partial observations.
Algorithm~\ref{alg: learning workflow} summarizes the loss composition during student training, with references to the detailed explanations in Section~\ref{subsec: model architecture} and \ref{subsec: learning objectives}.

\begin{algorithm}[!htb]
\caption{Student training in SenseAlign}
\label{alg: learning workflow}
\begin{algorithmic}[1]
\Require Pretrained teacher model
\State Randomly initialize the student model
\While{Training budget remains}
    \State Run current student policy in the environment and collect rollouts
    \State Store the corresponding future $\mathrm{GT}$ alongside the standard PPO rollout data.
    \For{Training Batch \textbf{in} Collected Rollouts}
    \Statex \hspace{3em}--- \textit{Prediction loss}
    \State Encode student observations $O$ \hfill Equation~\ref{eq: input embedding} and \ref{eq: road encoder}
    \State Predict future traffic with the student's predictor \hfill Equation~\ref{eq: prediction}
    \State Compute prediction loss $\Lpred$ \hfill Equation~\ref{eq: prediction loss}

    \Statex \hspace{3em}--- \textit{PPO loss}
    \State Aggregate student features from road nodes to grid cells \hfill Equation~\ref{eq: cell features} and \ref{eq: agent encoding}
    \State Compute actions and critic values for student \hfill Equation~\ref{eq: joint action decoding} and \ref{eq: value estimation}
    \State Compute the standard PPO loss $\Lppo$ from the collected rollout data \hfill Equation~\ref{equ: ppo final loss}
    \Statex \hspace{3em}--- \textit{Knowledge distillation loss}
    \State Encode teacher observations with privileged future GT information \hfill Equation~\ref{eq: future embedding} $\sim$ \ref{eq: teacher fusion}
    \State Aggregate GT-conditioned teacher features to grid cells \hfill Equation~\ref{eq: cell features} and \ref{eq: agent encoding}
    \State Compute Teacher($O$ | GT), the GT-conditioned teacher action on current state \hfill Equation~\ref{eq: joint action decoding}
    \State Compute knowledge distillation loss $\Lkd$ \hfill Equation~\ref{eq: action distillation}
    \Statex \hspace{3em}--- \textit{Decision alignment loss}
    \State Encode teacher observations while replacing future GT with student's Pred \hfill Equation~\ref{eq: future embedding} $\sim$ \ref{eq: teacher fusion}
    \State Aggregate Pred-conditioned teacher features to grid cells \hfill Equation~\ref{eq: cell features} and \ref{eq: agent encoding}
    \State Compute Teacher($O$ | Pred), the teacher's action distributions using $\mathrm{Pred}$ in place of $\mathrm{GT}$ \hfill Equation~\ref{eq: joint action decoding}
    \State Compute decision alignment loss $\Ldec$ \hfill Equation~\ref{eq: decision alignment}

    \Statex \hspace{3em}--- \textit{Student update}
    \State Combine PPO and auxiliary losses into the final loss $L$ \hfill Equation~\ref{eq: student loss}
    \State Update the student encoder, actor, critic, and predictor by minimizing $L$.
    \EndFor
\EndWhile
\end{algorithmic}
\end{algorithm}

\paragraph{Deployment}
As shown in Figure~\ref{fig: system}(c), when testing the student model, only the shared feature encoder and the actor are needed.
The predictor, critic, and teacher are present during training but not required for deployment.
Therefore, our learning framework will not introduce any test-time cost.

\subsection{Model architecture}
\label{subsec: model architecture}
\begin{figure}[!htb]
\centering
\includegraphics[width=\linewidth]{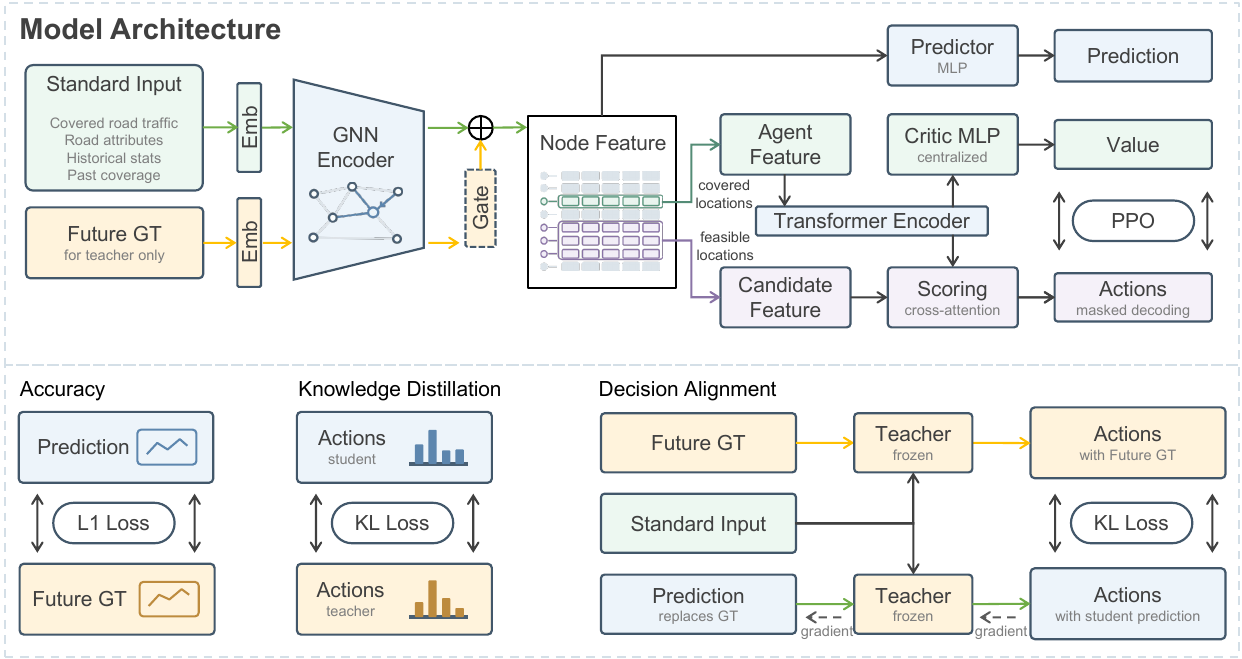}
\caption{
Model architecture (top) and auxiliary training objectives (bottom).
A shared road encoder supports action selection, value estimation, and traffic prediction.
The teacher additionally fuses privileged future traffic through a learned gate.
The three auxiliary objectives supervise traffic predictions, distill the teacher's actions into the student policy and align the prediction-induced teacher decisions.}
\label{fig: model}
\end{figure}

Figure~\ref{fig: model} demonstrates the details of model architecture and auxiliary training objectives.
The teacher and student share the same architecture, with the only difference being the teacher-only privileged future information branch.
Although SenseAlign's two-stage learning framework has no restrictions on the model design, we share the same structure between the student and teacher to quantify the value of privileged information.
The model starts with a GNN encoder to learn the network dynamics, and the prediction branch directly uses the produced feature to forecast next-step traffic.
an attention-based agent-target interaction module for actions and values, as well as the prediction module to enhance the representation space.

The remainder of this section describes the information flow from input data to decisions for one step and omits the time index for simplicity.
Throughout this section, we use $\mathbf x$ as a notation for input data, $\mathbf e$ for embeddings, and $\mathbf h$ for hidden features, with uppercase symbols denoting matrices of stacked vectors.
Superscripts identify the feature type or processing stage, while subscripts $i$, $j$, and $z$ index roads, drones and cells, respectively.

\paragraph{Fundamental Building Blocks}
Our model organizes a few fundamental neural-network operations to transform traffic measurements into features for prediction and decision-making.
Here we briefly introduce them and use $\mathbf{x}$ as a general notation for the inputs (raw data or hidden features).
A linear layer applies an affine transformation to its input data:
\begin{equation}
    \operatorname{Linear}(\mathbf x)=\mathbf W\mathbf x+\mathbf b,
\end{equation}
where the weight matrix $\mathbf W$ and bias vector $\mathbf b$ are learned during training.
Notably, an embedding layer (Embed) is a Linear layer mapping raw input data to hidden features.
Query and Key are also Linear layers in Transformer-style \citep{vaswani2017attention} attention scores.
The rectified linear unit (ReLU) sets the negative elements to zero and preserve the positive ones:
\begin{equation}
    \operatorname{ReLU}(\mathbf x)=\max(\mathbf x,\mathbf 0).
\end{equation}
LayerNorm normalizes the feature each feature vector using the mean and variance across its channels, followed by a learned scale and shift \citep{ba2016layernorm}:
\begin{equation}
\operatorname{LayerNorm}(\mathbf x)
=\frac{\mathbf x-\E[\mathbf x]}{\sqrt{\operatorname{Var}[\mathbf x]+\epsilon}}
    \odot\gamma+\beta,
\label{eq: layer normalization}
\end{equation}
where $\E[\mathbf x]$ and $\operatorname{Var}[\mathbf x]$ are the mean and variance across feature channels, $\gamma$ and $\beta$ are learned scale and shift vectors, and $\odot$ denotes elementwise multiplication.
The small positive constant $\epsilon$ ensures numerical stability when the variance is close to zero.
A multilayer perceptron (MLP) stacks multiple linear layers with nonlinear activations.
For example, an MLP with one hidden layer computes
\begin{equation}
    \operatorname{MLP}(\mathbf x)=\operatorname{Linear}(\operatorname{ReLU}(\operatorname{Linear}(\mathbf x))),
\end{equation}
where the two linear layers have independent parameters.
The Graph Convolution (GCNConv) exchanges information between road features and those of their neighbors, and the elementwise transform follows \citep{kipf2017gcn}:
\begin{equation}
\mathbf x'_i
=
\operatorname{Linear}\left( 
    \sum_{k\in\mathcal N(i)}
    \frac{e_{k,i}}{\sqrt{\hat d_k\hat d_i}}\mathbf x_k 
\right), \quad \text{where} \quad
\hat d_i=1+\sum_{k\in\mathcal N(i)}e_{k,i}.
\label{eq: graph convolution}
\end{equation}
Here, $\mathbf x_i$ and $\mathbf x'_i$ are features of road $i$ before and after GCNConv, and $\mathcal N(i)$ denotes its graph neighbors including itself.
The scalar $e_{k,i}$ is the weight of the edge from road $k$ to road $i$, which is uniformly set to one.
In this work, we use a topology-based graph, where the roads are graph neighbors if they are directly connected in the physical road network.

\paragraph{GNN-based Node Feature Encoder}
The encoder consumes raw data inputs and produces shared hidden features for subsequent neural network modules.
In the inputs, the raw features combine available traffic measurements with observation history and road attributes, allowing the encoder to account for both static and dynamic information.
For each road, the observation vector $\mathbf x_i^{\mathrm{obs}}\in\mathbb R^8$ contains the following feature channels:
\begin{itemize}
    \item Currently observed flow, density, and current time of day
    \item Last-observed flow and density, and observation age (time since last observation)
    \item Binary coverage mask indicating whether a road is currently covered by any drone
    \item Binary observation flag indicating whether a road is ever observed in the monitoring session
\end{itemize}
The historical vector $\mathbf x_i^{\mathrm{hist}}\in\mathbb R^4$ contains the training mean and standard deviation of flow and density for that road and time of day.
The same data normalization is applied to current, last-observed and historical flow and density as they have the same range and physical units.
Besides, we use the length and importance score in Equation~\ref{eq: road weight} of the road as its attribute vector $\mathbf x_i^{\mathrm{attr}}\in\mathbb R^2$, normalized by the mean over all roads respectively.

After data normalization, the observation, historical, and attribute vectors are embedded into $\mathbf{e}_i\in\mathbb R^D$:
\begin{equation}
\mathbf{e}_i=\operatorname{Embed}(\mathbf x_i^{\mathrm{obs}})
    +\operatorname{Embed}(\mathbf x_i^{\mathrm{hist}})
    +\operatorname{Embed}(\mathbf x_i^{\mathrm{attr}}),
\label{eq: input embedding}
\end{equation}
where the embedding layers are implemented with separate Linear layers with output dimension $D$.
Combining all the per-road $\mathbf{e}_i$ across roads gives the overall embedding $\mathbf E \in \mathbb R^{N \times D}$.
A learnable road position encoding $\mathbf E^{\mathrm{id}} \in \mathbb R^{N \times D}$ is then combined with the embeddings to provide location information.
The GNN encoder mainly consists of graph convolutional network (GCN) layers \citep{kipf2017gcn} for exchanging messages across connected roads.
We use skip connections to allow easy preservation of initial states.
\begin{equation}
\begin{aligned}
    \mathbf H^{\mathrm{local}}&=\operatorname{LayerNorm}(\mathbf E+\mathbf E^{\mathrm{id}}),\\
    \mathbf H^{(0)}&=\operatorname{Linear}(\mathbf H^{\mathrm{local}}),\\
    \mathbf H^{(1)}&=\operatorname{ReLU}\left(\operatorname{LayerNorm}
         (\operatorname{GCNConv}(\mathbf H^{(0)},G))\right),\\
    \mathbf H^{(2)}&=\operatorname{ReLU}\left(\operatorname{LayerNorm}
         (\operatorname{GCNConv}(\mathbf H^{(1)},G))\right),\\
    \mathbf H^{(3)}&=\operatorname{Linear}\left(
         \operatorname{GCNConv}(\mathbf H^{(0)}+\mathbf H^{(2)},G)\right),\\
    \mathbf H^{\mathrm{road}}&=\operatorname{Concat}(\mathbf H^{\mathrm{local}},\mathbf H^{(3)}).
\end{aligned}
\label{eq: road encoder}
\end{equation}
The forward pass keeps the hidden dimension unchanged until the final $\operatorname{Concat}$ step, which concatenates the local features $\mathbf H^{\mathrm{local}}$ with the graph features $\mathbf H^{(3)}$.
In fact, the last step doubles the channel width for each road to preserve initial local information alongside spatial context, resulting in a feature $\mathbf{H}^{\mathrm{road}} \in \mathbb R^{N \times 2D}$

\paragraph{Prediction branch}
The student's prediction branch reads the shared road features directly and uses an MLP to estimate normalized next-step flow and density for every road:
\begin{equation}
\mathrm{Pred}=\operatorname{MLP}(\mathbf H^{\mathrm{road}}).
\label{eq: prediction}
\end{equation}
Because the actor, critic, and predictor share $\mathbf H^{\mathrm{road}}$, prediction supervision also shapes the representation used to choose the sensing locations.

\paragraph{Privileged traffic fusion}
The teacher model has the same architecture, but adds a branch to encode future traffic information, similar to the normal observation embedding in Equation \ref{eq: input embedding}.
\begin{equation}
\mathbf e_i^{\mathrm{future}}
=\operatorname{Embed}(\mathbf x_i^{\mathrm{future}})
+\operatorname{Embed}(\mathbf x_i^{\mathrm{hist}})
+\operatorname{Embed}(\mathbf x_i^{\mathrm{attr}}),
\label{eq: future embedding}
\end{equation}
where $\mathbf x_i^{\mathrm{future}}$ is the future flow and density of road $i$, embedded also with a linear layer. 
Meanwhile, the historical and attribute features are reused from the normal observation branch.
Then, the future embeddings of all roads $\mathbf E^{\mathrm{future}} \in \mathbb R^{N \times D}$ are given to the shared spatial encoder as in Equation~\ref{eq: road encoder}.
\begin{equation}
\mathbf H^{\mathrm{future}}=\operatorname{SpatialEncoder}(\mathbf E^{\mathrm{future}},G).
\label{eq: teacher encoding}
\end{equation}
The gate module in Figure~\ref{fig: model} then fuses the outputs from the two branches as:
\begin{equation}
\mathbf H^{\mathrm{teacher}}=\mathbf H^{\mathrm{obs}}
    +\alpha\,\operatorname{Linear}\left(
        \operatorname{Concat}\left(
            \mathbf H^{\mathrm{obs}}, \mathbf H^{\mathrm{future}}
        \right)
    \right),
\label{eq: teacher fusion}
\end{equation}
where $\mathbf H^{\mathrm{obs}} \in \mathbb R^{N \times 2D}$ denotes the hidden states from the normal branch, and a learnable scalar parameter $\alpha$ controls the contribution of future traffic.
In fact, the Linear layer compresses the concatenated features back to $2D$ dimension.
The grid feature aggregation in the teacher model will use the fused $\mathbf H^{\mathrm{teacher}} \in \mathbb R^{N \times 2D}$.

\paragraph{Grid Feature Aggregation}
Since the monitoring agent operates on grid cells, the road features are aggregated by grids that drones can visit.
Let $I_z=\{i:g(i)=z\}$ denote the set of roads in cell $z$, the cell's attribute vector is aggregated from the roads within the cell:
\begin{equation}
\mathbf x_z^{\mathrm{attr}} =
\log \left(
    1+
    \operatorname{Concat}\left(
    |I_z|, \sum_{i\in I_z}\mathbf x_i^{\mathrm{attr}}
    \right) 
\right),
\end{equation}
where the $\log$ is applied elementwise.
For the cell feature $\mathbf h_z^{\mathrm{cell}} \in \mathbb R^{2D}$, we apply average pooling over the road features in the cell, and add it to the projected cell attribute:
\begin{equation}
\mathbf h_z^{\mathrm{cell}} =
\operatorname{Mean}_{i\in I_z}(\mathbf h_i^{\mathrm{road}}) + 
\operatorname{Linear} (\mathbf x_z^{\mathrm{attr}}).
\label{eq: cell features}
\end{equation}
With the cell features, each drone combines its current cell feature with the cell's positional encoding $\mathbf e_{z_j}^{\mathrm{pos}} \in \mathbb R^{D/2}$ (a set of learnable parameters) and a global context obtained by averaging all road features in the whole road network:
\begin{equation}
\begin{aligned}
    \mathbf h^{\mathrm{global}}&=\operatorname{ReLU}\left(
       \operatorname{Linear}(\operatorname{Mean}_i(\mathbf h_i^{\mathrm{road}}))
    \right),\\
    \mathbf h_j^{\mathrm{agent}}&=\operatorname{ReLU}\left(
        \operatorname{Linear}\left(
            \operatorname{Concat}\left(
                \mathbf h_{z_j}^{\mathrm{cell}},\mathbf e_{z_j}^{\mathrm{pos}},\mathbf h^{\mathrm{global}}
            \right)
        \right)
    \right).
\end{aligned}
\label{eq: agent encoding}
\end{equation}
Both Linear layer here have output dimension $D'$ (different from the feature encoder).
As a result, the agent feature is a joint representation containing traffic observation, grid location and network context.
Then, a Transformer encoder exchanges information among the agents, producing a set of attended agent features $\widetilde{\mathbf h}_j^{\mathrm{agent}} \in \mathbb R^{D'}$.

\paragraph{Actor and Critic}
Following the common practice in RL, the actor is trained to predict preference scores over a set of possible candidate destination cells, which are non-empty cells within grid distance $\rho=3$.
For this purpose, we utilize cross-attention between the attended agent feature $\widetilde{\mathbf h}_j^{\mathrm{agent}}$ and the feature of the candidate destination $\mathbf h_z^{\mathrm{cell}}$.
For a drone $j$ and a candidate displacement leading to a candidate cell $z$, a preference score is given by the following formula:
\begin{equation}
    s_{j,z}=
    \frac{\operatorname{Query}(\widetilde{\mathbf h}_j^{\mathrm{agent}})^\top
      \operatorname{Key}(\mathbf h_{z}^{\mathrm{cell}})}{\sqrt d}
      + b,
    \label{eq: destination score}
\end{equation}
where Query($\cdot$) and Key($\cdot$) are linear layers with output dimensions $D'$, and $b$ is a learnable bias parameter.
The query represents the drone's sensing context, and the key represents the status of the candidate cells, thus their compatibility determines the preference for each destination.

Given the preference scores, the drone movements are decoded autoregressively from the first to the last according to the drone index $1,\ldots,M$.
The purpose is to allow later drones to account for locations chosen by previous ones and avoid conflicts.
Concretely, a drone is not allowed to move to an empty grid cell with no roads or a cell occupied by other drones.
Such disallowed actions will be marked as invalid (equivalent to setting the scores to $-\infty$), and a SoftMax over the remaining scores gives the conditional action distribution.
Therefore, the joint policy factorizes into:
\begin{equation}
    \pi(\mathbf{u}\mid O)
       =\prod_{j=1}^{M}\pi(\mathbf{u}_j\mid O,\mathbf{u}_{<j}),
    \label{eq: joint action decoding}
\end{equation}
where $\mathbf{u}_{<j}$ denotes actions already chosen in the current step.
In training, the policy samples from the masked conditional distribution for balanced exploration and exploitation, and during evaluation, the policy selects the available action with the highest score.

The critic or value network estimates the expected return from a state if the agent follows the learned policy.
Therefore, we implement a centralized critic with MLP using global feature and the averaged agent feature as inputs:
\begin{equation}
    V(O) = \operatorname{MLP}\left(
        \operatorname{Concat}\left(
            \mathbf h^{\mathrm{global}}, 
            \operatorname{Mean}_j(\mathbf h_j^{\mathrm{agent}})
        \right)
    \right)
\label{eq: value estimation}
\end{equation}

\subsection{Learning objectives}
\label{subsec: learning objectives}
The lower part of Figure~\ref{fig: model} shows how the auxiliary losses are connected with the model, in addition to PPO \citep{schulman2017ppo}.
These losses provide supervision signals on various network outputs, but all of them are based on the same shared feature encoder, therefore shaping a better latent space for decision-making together.

\paragraph{Accuracy}
In the student model, the predictor branch is trained to predict future traffic for the complete road network.
This branch is trained to minimize the mean absolute error (MAE or L1 loss) with respect to the future ground-truth traffic:
\begin{equation}
    \begin{aligned}
    \Lpred = \operatorname{MAE}(\mathrm{Pred}, \mathrm{GT}) = \frac{1}{NC}\sum_{i=1}^{NC}\left\|\hat{\mathbf y}_i-\mathbf y_i\right\|_1,
    \end{aligned}
\label{eq: prediction loss}
\end{equation}
where $\mathbf y$ and $\hat{\mathbf y}$ are flattened ground-truth and predicted future traffic states (flow and density), each with $N \times C$ elements.
This objective encourages the encoder to infer how traffic evolves over the whole road network.

\paragraph{Knowledge distillation}
During the process of RL training, the student model may not be able to fully explore the environment due to its imperfect observations and may converge early to local minima.
Therefore, we introduce a privileged teacher model to guide the student agent's decision-making and help the student learn better than it would with the reward signal alone.
Specifically, we use the Kullback-Leibler (KL) divergence as the loss for Knowledge Distillation (KD):
\begin{equation}
\begin{aligned}
    \Lkd &= \KL\left(\StopGrad\left[\Teacher(\cdot,\mathrm{GT})\right]\;\Vert\;\Student(\cdot)\right) \\
    &= \sum_{j=1}^{M}\sum_{\mathbf u_j\in U_j}p_{T,j}(\mathbf u_j)
       \log\frac{p_{T,j}(\mathbf u_j)}{p_{S,j}(\mathbf u_j)},
\end{aligned}
\label{eq: action distillation}
\end{equation}
where $\mathbf u_j$ is the movement action of drone $j$ in the valid action set $U_j$.
$p_{T,j}$ and $p_{S,j}$ are the GT-conditioned teacher and student action distributions for drone $j$, both computed from the collision-aware decoding in Equation~\ref{eq: joint action decoding}.
During this process, the conditional action probabilities are decoded according to the recorded actions in the rollout buffer collected by the old student model, rather than the high-score actions from the teacher and student models being trained.
That implementation allows the KL divergence to be computed on a consistent set of actions and avoid numerical instability.
$\StopGrad$ is an operator to stop gradient flow in backpropagation, which holds the teacher model frozen.
At the same time, the student model is allowed to update.
To summarize, knowledge distillation trains the student to imitate the teacher's preferred actions even if the student only has partial observations itself.

\paragraph{Decision alignment}
The purpose of the prediction branch is to learn better representations with knowledge of system dynamics.
However, prediction accuracy doesn't necessarily align with the ultimate goal of decision-making.
We draw inspiration from the SPO theory \citep{elmachtoub2022smart}, which proposes that a predictor should minimize the decision deviation from the optimal solution, rather than some prediction error.
In other words, the prediction quality is evaluated by the downstream decision-making task.
Therefore, we propose to align the prediction with decisions using a decision alignment objective.
We train the prediction branch so that when its prediction replaces the future ground-truth in the input of a frozen teacher model, the teacher's actions remain similar:
\begin{equation}
\begin{aligned}
    \Ldec=\KL\left(\StopGrad[\Teacher(\cdot,\mathrm{GT})] \; \Vert \; \Teacher^*(\cdot,\mathrm{Pred})\right).
\end{aligned}
\label{eq: decision alignment}
\end{equation}
As in Equation~\ref{eq: action distillation}, we entirely stop the gradient of the GT-conditioned branch to keep the teacher from updating.
Additionally, the operator $\mathrm{Teacher^*}(\cdot)$ means the teacher's parameters are frozen, but the student's prediction remains differentiable in the computation graph.
This small difference allows the loss to pass through the teacher and reach the predictions; thus the predictor and feature backbone can be updated in backpropagation.
In summary, the teacher model provides a decision-specific measurement of forecast quality, and the student's predictor is trained to preserve decision-critical information.

\paragraph{Policy optimization}
The monitoring agent is optimized following the general workflow of PPO \citep{schulman2017ppo} implemented in Stable-Baselines~3 \citep{stable-baselines3}, with three auxiliary learning objectives described above.
Concretely, we use the clipped policy loss as follows:
\begin{equation}
L_{\mathrm{clip}} =-\mathbb E_{O,\mathbf{u}\sim\pi_{\mathrm{old}}}\left[
\min\left(r A,\operatorname{clip}\left(r,1-\epsilon,1+\epsilon\right)A\right)\right],
\quad 
r = \frac{\pi(\mathbf{u}\mid O)}{\pi_{\mathrm{old}}(\mathbf{u}\mid O)}
\label{eq: ppo policy loss}
\end{equation}
where $\pi_{\mathrm{old}}$ is the policy in the previous training round, $A$ is the normalized advantage estimate, and $\epsilon$ is the clipping threshold.
The advantage is a transformed reward indicating how much better the action is compared to the average case, and it is computed with generalized advantage estimation (GAE):
\begin{equation}
\hat A_t = \sum_{\ell=0}^{T-t-1}(\gamma\lambda)^\ell\delta_{t+\ell}, 
\quad \text{where} \quad
\delta_t = r_t+\gamma V_{\mathrm{old}}(O_{t+1})-V_{\mathrm{old}}(O_t)
\label{eq: gae}
\end{equation}
where $V_{\mathrm{old}}$ is the critic used during rollout collection (from the previous round), and $\delta_t$ is the temporal-difference residual.
$\gamma$ and $\lambda$ together control the decay of future residuals.
Meanwhile, the critic is trained to fit the return target:
\begin{equation}
L_V = \mathbb E_t\left[(V(O_t)-\hat R_t)^2\right],
\quad \text{where} \quad
\hat R_t = V_{\mathrm{old}}(O_t)+\hat A_t
\label{eq: ppo value loss}
\end{equation}
where $\hat R_t$ is the fixed return target computed from the unnormalized GAE estimate, and $V$ is the critic being optimized.
The loss of PPO is a weighted sum of policy and value losses, as well as an entropy term to encourage exploration:
\begin{equation}
\Lppo = L_{\mathrm{clip}}+c_VL_V-c_HH.
\label{equ: ppo final loss}
\end{equation}
The entropy $H$ is summed over the per-drone action distribution entropy, while $c_V$ and $c_H$ control the weights of value regression and entropy regularization.
Finally, we combine the standard PPO loss with our auxiliary losses to build the student learning objective as follows:
\begin{equation}
    L=\Lppo
       +c_{\mathrm{pred}}\Lpred
       +c_{\mathrm{KD}}\Lkd
       +c_{\mathrm{dec}}\Ldec.
    \label{eq: student loss}
\end{equation}
where the coefficients $c_{\mathrm{pred}}$, $c_{\mathrm{KD}}$ and $c_{\mathrm{dec}}$ weight the three auxiliary losses respectively.

\section{Experiments}
\label{sec: experiments}

This section will describe the experimental setup (Section~\ref{subsec: experiment setup}), compare overall performance (Section~\ref{subsec: overall performance}), provide ablations and sensitivity analysis (Section~\ref{subsec: ablation and sensitivity analysis}), and present detailed hyperparameters and training records (Section~\ref{subsec: additional implementation and training details}).

\subsection{Experiment setup}
\label{subsec: experiment setup}

\begin{figure}[!htb]
\centering
\begin{subfigure}[c]{0.37\linewidth}
\centering
\includegraphics[width=\linewidth]{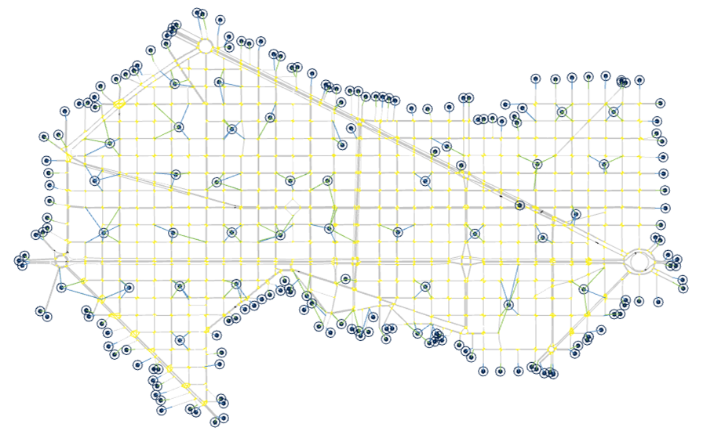}
\caption{Road network}
\end{subfigure}\hfill
\begin{subfigure}[c]{0.37\linewidth}
\centering
\includegraphics[width=\linewidth]{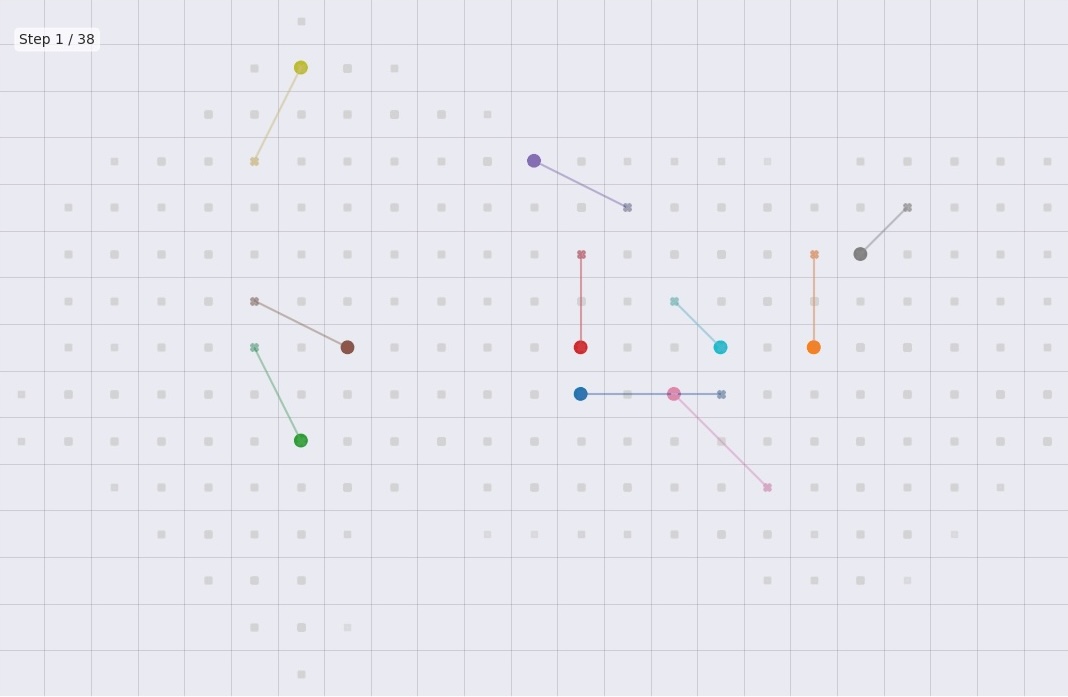}
\caption{Monitoring grid with 10 drones}
\end{subfigure}\hfill
\begin{subfigure}[c]{0.25\linewidth}
\centering
\includegraphics[width=\linewidth]{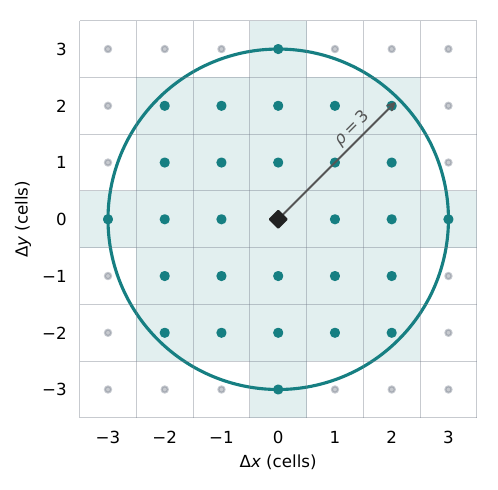}
\caption{Reachable cells ($\rho=3$)}
\label{fig: movement offsets}
\end{subfigure}
\caption{Active urban sensing setup.
Roads in the road network are divided into cells, and each drone observes the roads in its occupied cell.
A $\odot$ symbol in the road network (a) indicates a source or target location of vehicles.
The drones move on the grid (b) within a distance constraint (c).}
\label{fig: monitoring setting}
\end{figure}

\paragraph{Environment and evaluation}
We build the traffic-monitoring environment using the SimBarca dataset \citep{xiong2025multi}, which provides simulated flow and density on the real road network of central Barcelona.
The dataset contains 75 training sessions and 26 evaluation sessions with different morning-peak travel demands.
We take the congestion onset stage of the sessions while omitting the warm-up and clearance, resulting in 39 time steps at three-minute intervals for each session.
As shown in Figure~\ref{fig: monitoring setting}, the network contains 1,570 road segments and is divided into grid cells of size $220\,\mathrm{m}\times220\,\mathrm{m}$.
We assume that a drone hovering over a cell center observes traffic within that cell without measurement error, with monitoring rewards computed from these observations.

At the beginning of each episode, drones are placed randomly in cells containing roads.
Their actions follow Equation~\ref{eq: movement}, with a movement radius of $\rho=3$ cells per step (Figure~\ref{fig: movement offsets}).
At a speed of $10\,\mathrm{m/s}$, the maximum movement takes approximately one minute within each three-minute sensing interval.
We assume that this travel time does not reduce observation quality or the reward obtained at the destination.
A drone remains in place if it attempts to leave the valid region, and drones attempting to occupy the same cell also remain in place.
For evaluation, we use random seed 888 to generate five initial-position configurations for each session and average the results over these configurations.

\paragraph{Baselines}
The PPO baseline uses the same architecture (Figure~\ref{fig: model}) and action decoder (Equation~\ref{eq: joint action decoding}) as SenseAlign, with all three auxiliary losses disabled.
We also include three non-learning baselines.
Static Stay keeps each drone at its initial position, Random Walk selects movements randomly from the feasible actions, and Sweeping divides a lawnmower route equally among the drones and has them follow a back-and-forth scan.
We further adapt two multi-agent policy learning methods, MAGPPO \citep{chen2021monitoring} and MAT \citep{wen2022mat}.
Following MAGPPO's image-based input, we aggregate road observations within spatial grids and encode the resulting map with a CNN.
Both adaptations use this observation extractor to form per-drone features.
MAGPPO coordinates drones through graph-attention communication, while MAT generates their actions autoregressively.

\paragraph{Training Hyperparameters}
Except for the fleet-size experiment, we use 10 drones, providing about 5\% spatial coverage.
For SenseAlign, we train a separate teacher for each monitoring task using 8M environment steps.
Student training then uses 8M steps for congestion-change monitoring and 4M steps for MFD estimation.
The auxiliary-loss coefficients in Equation~\ref{eq: student loss} are $c_{\mathrm{pred}}=10^{-4}$, $c_{\mathrm{KD}}=0.1$, and $c_{\mathrm{dec}}=0.1$.
PPO uses the same training settings and task-specific budgets as the SenseAlign student.
Following Stable-Baselines3 \citep{stable-baselines3}, both methods sample from the action distribution during training and select the highest-probability action during evaluation.
We evaluate the final checkpoint and report total episode reward for congestion-change monitoring and mean estimation error for MFD (Equation~\ref{eq: mfd error}).
Each training experiment is repeated with three random seeds, and we report the mean and sample standard deviation (SD) across seeds.

\subsection{Overall performance}
\label{subsec: overall performance}

\begin{table}[!htb]
\centering
\caption{Overall monitoring performance, reported as mean $\pm$ sample SD}
\label{tab: main}
\begin{tabular}{lcc}
\toprule
Method & Congestion change $\uparrow$ & MFD error (\%) $\downarrow$ \\
\midrule
Static Stay & $4.07 \pm 0.19$ & $16.44 \pm 1.08$ \\
Random walk & $11.92 \pm 0.56$ & $16.46 \pm 0.21$ \\
Sweeping & $14.22 \pm 1.25$ & $18.40 \pm 0.44$ \\
MAT \citep{wen2022mat}  & $14.17 \pm 0.32$ & $14.40 \pm 0.19$ \\
MAGPPO \citep{chen2021monitoring}  & $18.11 \pm 0.34$ & $12.79 \pm 0.29$ \\
\midrule
PPO \citep{schulman2017ppo} & 28.85 $\pm$ 0.46 & 6.43 $\pm$ 0.25 \\
SenseAlign (ours) & 31.20 $\pm$ 0.45 & 5.70 $\pm$ 0.27 \\
\midrule
Privileged teacher & 35.57 $\pm$ 1.47 & 5.08 $\pm$ 0.32 \\
\bottomrule
\end{tabular}
\end{table}

Table~\ref{tab: main} compares SenseAlign with the baseline methods on both monitoring tasks, where  SenseAlign achieves the best performance among the methods using partial observations.
For congestion change monitoring, the privileged teacher achieves $35.57\pm1.47$ due to its advantage of knowing future traffic, showing the value of future information in active sensing.
The SenseAlign student model scores $31.20\pm0.45$ under a realistic 5\% spatial coverage, which improves over PPO's $28.85\pm0.46$ by 8.1\%.
The other baselines generally have lower metrics than PPO, and MAGPPO has the highest average congestion change score at 18.11 among them.
MFD estimation shows a similar performance hierarchy; SenseAlign is the best method with partial observation, followed by PPO and the remaining baselines.
SenseAlign reduces the MFD error from PPO's 6.43\% to 5.70\%, which is a relative improvement of $\sim$11\%.
The performance gains show SenseAlign's framework supports coordination under distributed and fleet-level reward signals.

\begin{figure}[!htb]
\centering
\includegraphics[width=0.5\linewidth]{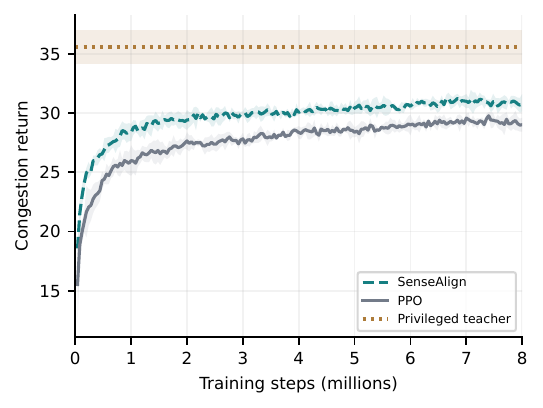}
\caption{Congestion-change learning curves.
Lines show mean returns and shaded bands indicate $\pm$ one sample SD across training seeds
}
\label{fig: reward curve}
\end{figure}

Figure~\ref{fig: reward curve} shows the training-time episode reward for congestion change monitoring, along with the performance of the trained teacher model.
Both SenseAlign and PPO learn steadily along the training process, and they saturate near the end.
After training, SenseAlign reaches higher rewards than PPO with a clear performance margin, but still has an obvious gap to the privileged teacher due to the realistic observation constraint.

\paragraph{Spatial allocation and monitoring value}
To study how the performance gains are related to sensing behavior, we compare spatial coverage and per-location reward contributions at five travel demand scales~\footnote{A demand scale is a multiplicative factor applied to the travel demand matrix, e.g., a demand scale of 1.8 means there are approximately 80\% more vehicles than under normal travel demand}.
For each cell, coverage is the fraction of time it is covered by the drone fleet, and reward contribution is the congestion-change return obtained from that cell.

\begin{figure}[!htb]
\centering
\includegraphics[width=\textwidth]{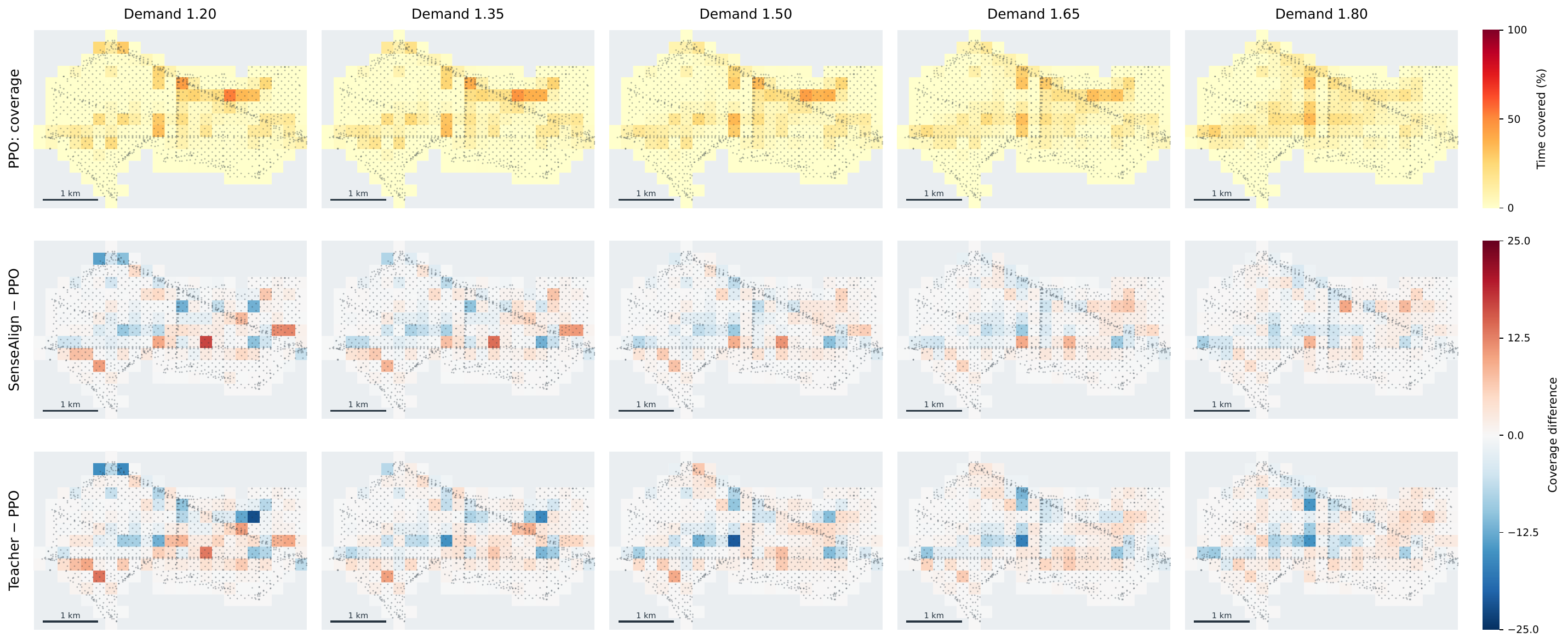}
\caption{Average spatial coverage for congestion change monitoring, grouped by traffic demand}
\label{fig: spatial coverage}
\end{figure}

\begin{figure}[!htb]
\centering
\includegraphics[width=\textwidth]{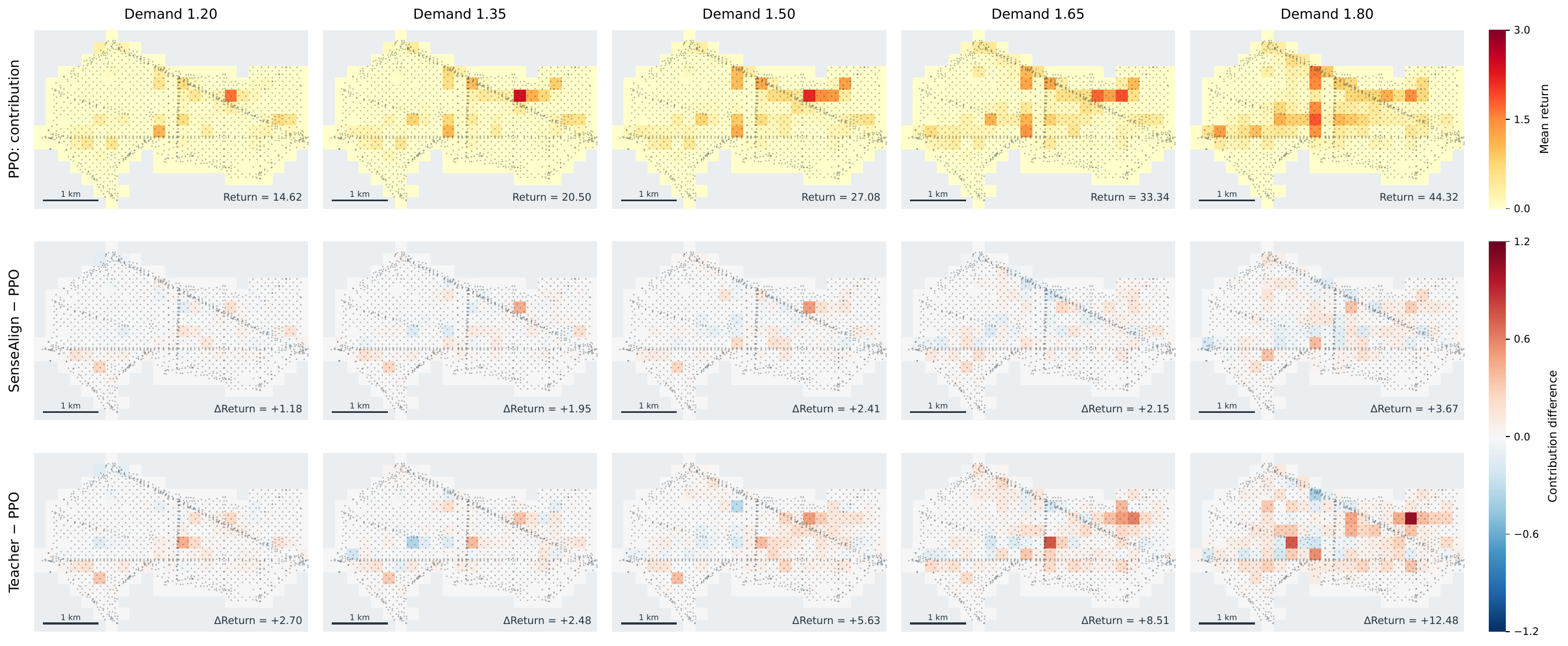}
\caption{Spatial reward contributions for congestion change monitoring, grouped by traffic demand}
\label{fig: spatial reward}
\end{figure}

The first rows of Figure~\ref{fig: spatial coverage} and Figure~\ref{fig: spatial reward} show the coverage and obtained reward by spatial location for PPO.
A clear pattern is that monitoring efforts are spent near the major roads, which is intuitively reasonable, as the traffic at those areas is more complex and worth monitoring.
Higher coverage also translates to higher rewards, meaning that the learned policy can allocate the resources according to the designed task.
As the travel demand increases, the drone coverage spreads more across the network, as the congestion becomes heavier, causes a wider impact on the road network and results in more congestion changes.

The other two rows of these two figures illustrate the difference from PPO for the teacher and SenseAlign respectively.
The coverage difference in Figure \ref{fig: spatial coverage} is usually less than 20\%, which indicates that SenseAlign navigates the drones also near major roads, similar to PPO.
The difference pattern is similar to that of the teacher model, showing that SenseAlign has learned from the teacher despite having only partial current observation.
Although the monitoring budget is fixed, Figure~\ref{fig: spatial reward} has shown that SenseAlign can visit the cells more intelligently, as the per-location return is higher than PPO's on most cells.
This pattern demonstrates that the overall performance gain is not obtained by exploiting a few locations, but by understanding the traffic evolution and optimizing drone allocations over space and time.

\begin{figure}[!htb]
\centering
\includegraphics[width=\textwidth]{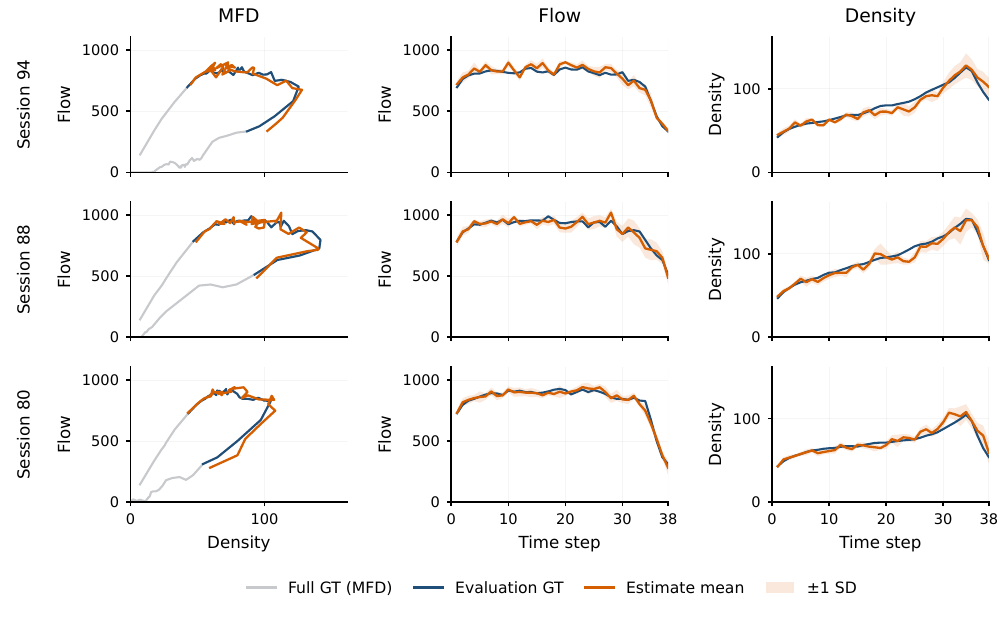}
\caption{MFD estimation quality for multiple simulation session. The units of flow and density are vehicle/hour and vehicle/km respectively.
The gray curve (Full GT) in the MFD plots presents the complete session including the network warmup and clearance stages, while the evaluations do not include those stages.}
\label{fig: mfd estimation}
\end{figure}

\paragraph{MFD Estimation Quality}
Figure~\ref{fig: mfd estimation} visualizes the quality of estimated MFD along with the corresponding flow and density components.
Although the experiments only use 10 drones covering 5\% of the road network, all of MFD, flow and density can be estimated with favorable quality. 
Five randomized evaluations are run for each of the training seeds, and the observed variance among the 15 runs are generally small, showing the stability of the learned monitoring agent.
Meanwhile, higher uncertainties are observed for density estimation at high-density scenarios, which reveals the challenge of identifying dynamic congestions.

\subsection{Ablation and Sensitivity Analysis}
\label{subsec: ablation and sensitivity analysis}

\paragraph{Ablation on auxiliary losses}
To quantify the effect of each proposed learning objective, we remove them from the complete SenseAlign and report the metrics in Table~\ref{tab: auxiliary ablation}, and PPO serves as the reference with all three auxiliary losses disabled.
In general, all three objectives contribute to the final performance of SenseAlign, as removing any of them results in a performance drop for both tasks.
All of them are helpful for the task as disabling one objective still results in higher performance than PPO.
The detailed results in Section \ref{subsec: additional implementation and training details} shows the improvement is consistent across random seeds.
Knowledge distillation turns out to be the most important factor, as its absence leads to the most degradation.

\begin{table}[!htb]
\centering
\caption{Auxiliary-loss ablations, reported as mean $\pm$ sample SD}
\label{tab: auxiliary ablation}
\begin{tabular}{lcc}
\toprule
Method & Congestion change $\uparrow$ & MFD error (\%) $\downarrow$ \\
\midrule
SenseAlign & 31.20 $\pm$ 0.45 & 5.70 $\pm$ 0.27 \\
w/o $\Lpred$ & 30.77 $\pm$ 0.45 & 5.82 $\pm$ 0.23 \\
w/o $\Ldec$ & 30.63 $\pm$ 0.40 & 5.77 $\pm$ 0.31 \\
w/o $\Lkd$ & 29.46 $\pm$ 0.71 & 6.18 $\pm$ 0.62 \\
PPO & 28.85 $\pm$ 0.46 & 6.43 $\pm$ 0.25 \\
\bottomrule
\end{tabular}
\end{table}

Table~\ref{tab: congestion demand} breaks down the ablation results further by traffic demand.
SenseAlign achieves the highest mean return in all five groups, and all learning objectives consistently bring improvements across all demand scales, showing the effectiveness of the proposed learning framework.
In addition, removing knowledge distillation produces the largest reduction among the single-loss ablations at every demand scale, consistent with the aggregate results in Table~\ref{tab: auxiliary ablation}.

\begin{table}[!htb]
\centering
\caption{Congestion-change return under auxiliary-loss ablations, grouped by traffic demand and reported as mean $\pm$ sample SD}
\label{tab: congestion demand}
\setlength{\tabcolsep}{3pt}
\begin{tabular}{lcccccc}
\toprule
Method & Demand 1.20 & Demand 1.35 & Demand 1.50 & Demand 1.65 & Demand 1.80 & Overall \\
\midrule
SenseAlign & $15.80 \pm 0.56$ & $22.45 \pm 0.66$ & $29.49 \pm 0.77$ & $35.49 \pm 0.84$ & $47.99 \pm 0.61$ & $31.20 \pm 0.45$ \\
w/o $\Lpred$ & $15.66 \pm 0.11$ & $22.32 \pm 0.46$ & $29.34 \pm 0.25$ & $35.24 \pm 1.18$ & $46.71 \pm 0.63$ & $30.77 \pm 0.45$ \\
w/o $\Ldec$ & $15.65 \pm 0.07$ & $22.03 \pm 0.65$ & $28.82 \pm 1.02$ & $34.93 \pm 0.73$ & $47.09 \pm 0.51$ & $30.63 \pm 0.40$ \\
w/o $\Lkd$ & $14.84 \pm 0.56$ & $21.40 \pm 0.83$ & $27.84 \pm 0.23$ & $34.58 \pm 1.05$ & $44.37 \pm 1.61$ & $29.46 \pm 0.71$ \\
PPO & $14.62 \pm 0.91$ & $20.50 \pm 0.73$ & $27.08 \pm 0.21$ & $33.34 \pm 1.41$ & $44.32 \pm 0.39$ & $28.85 \pm 0.46$ \\
\bottomrule
\end{tabular}
\end{table}

\begin{figure}[!htb]
\centering
\begin{subfigure}[t]{0.3\linewidth}
\centering
\includegraphics[width=\linewidth]{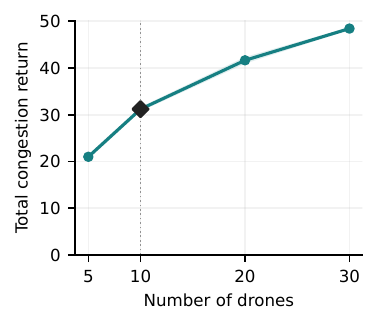}
\caption{}
\label{fig: drone scaling total}
\end{subfigure}
\begin{subfigure}[t]{0.3\linewidth}
\centering
\includegraphics[width=\linewidth]{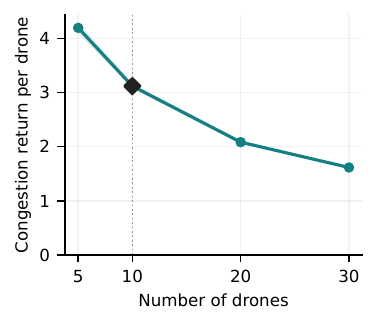}
\caption{}
\label{fig: drone scaling per drone}
\end{subfigure}
\caption{Congestion-change return by fleet size. Default values marked by black diamonds.}
\label{fig: drone scaling}
\end{figure}

\paragraph{Fleet size}
Intuitively, spending more resources for active sensing can improve the monitoring quality.
To quantify this effect, we experiment with congestion-change monitoring using SenseAlign with 5, 10, 20, and 30 drones (Figure~\ref{fig: drone scaling}).
The figure shows that when the fleet becomes larger, the overall performance improves but the per-drone return diminishes.
This is reasonable since a few drones can already cover the most valuable locations, and additional drones will have diminishing returns.
We find 10 drones to be a balanced size and use it as the default choice in all other experiments.

\begin{figure}[!htb]
\centering
\includegraphics[width=0.8\linewidth]{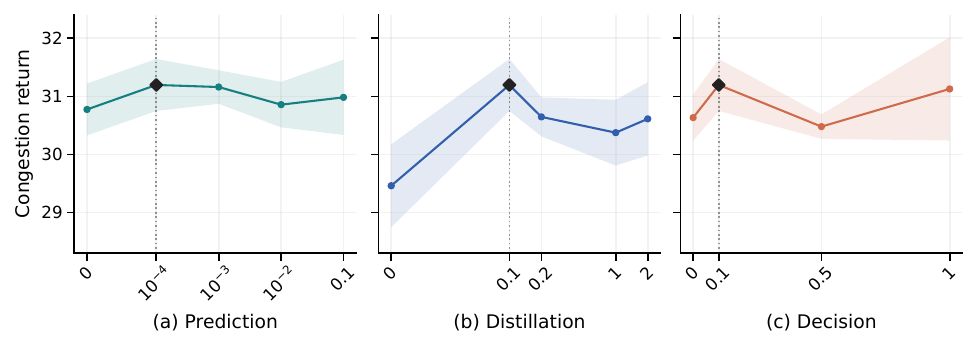}
\caption{Sensitivity to auxiliary-loss coefficients, shown as mean $\pm$ one sample SD.
Black diamonds mark the default coefficients.
The coefficients are: (a) prediction $\Lpred$, (b) knowledge distillation $\Lkd$ and (c) decision-alignment $\Ldec$}
\label{fig: coefficients}
\end{figure}

\paragraph{Loss coefficients}
Figure~\ref{fig: coefficients} examines the auxiliary-loss coefficients by varying one at a time and fixing the other two at their default values marked by the black diamond.
Unless an objective is completely disabled (set to 0), SenseAlign's performance remains stable across the examined coefficient range.
Specifically, we found setting the prediction coefficient at a smaller magnitude results in better overall performance.
This phenomenon indicates that the prediction objective is helpful for learning the traffic dynamics, but may not be well-aligned with the decisions.
Although the privileged teacher has higher rewards, simply increasing the knowledge distillation term won't make the student model better.
This is because the student can't reliably replicate the teacher's behavior with partial information.

\paragraph{Decision information in forecasts}
Finally, Table~\ref{tab: teacher input} presents a quantified experiment to demonstrate the effect of the decision-alignment objective.
Specifically, at each decision step, the trained teacher's future-traffic input is replaced by the predictions of the complete SenseAlign student model, the student predictions without decision-aligned learning, or the historical average per time-of-day.
Action agreement measures how often each drone selects the same action as the GT-informed teacher, and reward reduction indicates the performance degradation as a percentage.
In fact, this experiment illustrates the differences between the two KL terms in Equation~\ref{eq: decision alignment} using agreements in actions and the final rewards.

\begin{table}[!htb]
\centering
\caption{Frozen-teacher performance with different future-traffic inputs, reported as mean $\pm$ SD}
\label{tab: teacher input}
\begin{tabular}{lccc}
\toprule
Teacher input & Action agreement (\%) & Reward & Reduction (\%) \\
\midrule
Future GT & 100.00 & 35.57 $\pm$ 1.47 & 0.00 \\
SenseAlign & 33.18 $\pm$ 1.02 & 30.19 $\pm$ 0.85 & $-15.08$ \\
w/o $\Ldec$ & 21.91 $\pm$ 2.26 & 23.20 $\pm$ 1.18 & $-34.69$ \\
Historical Avg. & 26.15 $\pm$ 1.75 & 28.62 $\pm$ 0.39 & $-19.47$ \\
\bottomrule
\end{tabular}
\end{table}

With SenseAlign predictions, the teacher model takes the same action as the GT-conditioned case $\sim$33\% of the time and the collected reward is $\sim$85\% of that in the ideal case.
However, without decision alignment, the actions match only for $\sim$22\% of cases, resulting in much lower obtained reward, which is even worse than that obtained using the historical average.
This comparison has shown that our decision-alignment objective can effectively guide the feature encoder and predictor to preserve decision-related information.

\subsection{Additional Implementation and Training Details}
\label{subsec: additional implementation and training details}

\paragraph{Model Configuration}
In the SenseAlign model, the hidden dimension of the feature encoder is set to 64, which applies to the output dimensions of embedding and linear layers in Equation~\ref{eq: road encoder}.  
Road inputs and attributes are embedded into a 64-dimensional space, and the spatial encoder finally concatenates local and graph-propagated features to produce 128 channels per road.
For the graph convolutions, we use one-hop road adjacency based on topology, i.e., the roads that are physically connected in the road network are regarded as adjacent nodes in the graph representation.
The agent and cell features in Equation~\ref{eq: agent encoding} have 256 dimensions, with the grid position embedding having 32 dimensions.
The Transformer encoder for agent coordination has 4 heads and a feed-forward width of 512, and the query and key projections in Equation~\ref{eq: destination score} have width 256.

\paragraph{Optimization Hyperparameters}
In the two-stage learning framework, both the teacher and student models are trained for 8 million steps on congestion change monitoring task and 4 million steps for MFD estimation.
We build the workflow with PyTorch and Stable-Baselines 3 \citep{stable-baselines3}, and follow their standard practice on normalizing observations and advantages.
During rollout collection, we use 8 parallel environments for faster collection, and keep a buffer size of 512.
The models are trained with learning rate $5\times10^{-5}$ for 3 epochs using batch size 256.
For PPO-specific parameters, we set target KL to 0.02, entropy regularization $c_H=0.002$, value loss weight $c_V=0.5$, discount factor $\gamma=0.99$ and GAE parameter $\lambda=0.95$.
The default values for the auxiliary losses in SenseAlign are prediction loss $\Lpred=1\times 10^{-4}$, distillation loss $\Lkd=0.1$ and decision loss $\Ldec=0.1$ respectively.
We also follow the standard practices of Stable-Baselines 3 \citep{stable-baselines3} on normalizing input data and advantage.

\begin{figure}[!htb]
\centering
\includegraphics[width=\linewidth]{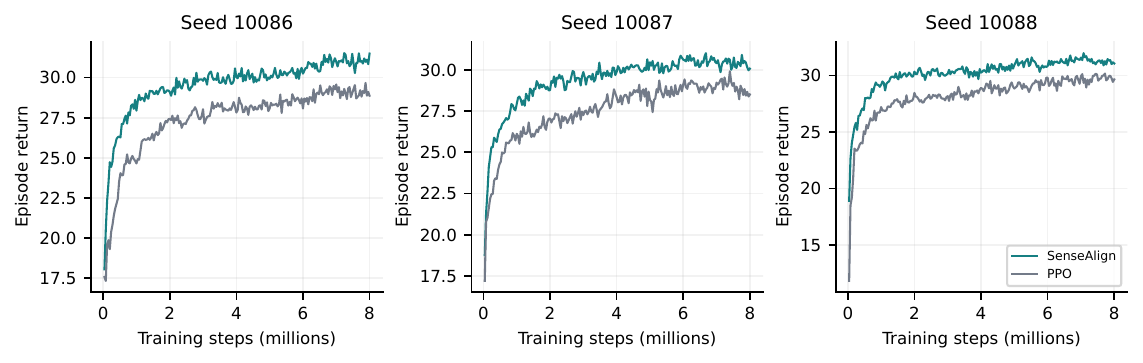}
\caption{The learning curves of congestion-change monitoring by training seed.}
\label{fig: reward curves all seeds}
\end{figure}

\begin{figure}[!htb]
\centering
\begin{subfigure}[t]{0.33\linewidth}
\centering
\includegraphics[width=\linewidth]{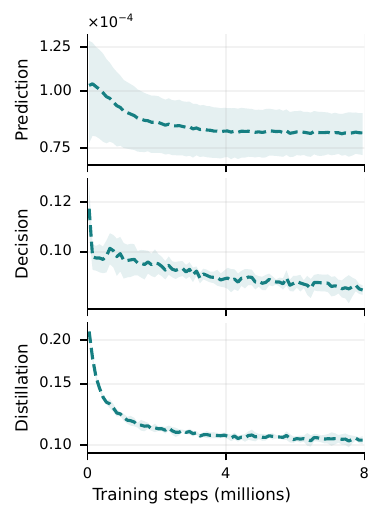}
\caption{SenseAlign auxiliary objectives}
\end{subfigure}\hfill
\begin{subfigure}[t]{0.66\linewidth}
\centering
\includegraphics[width=\linewidth]{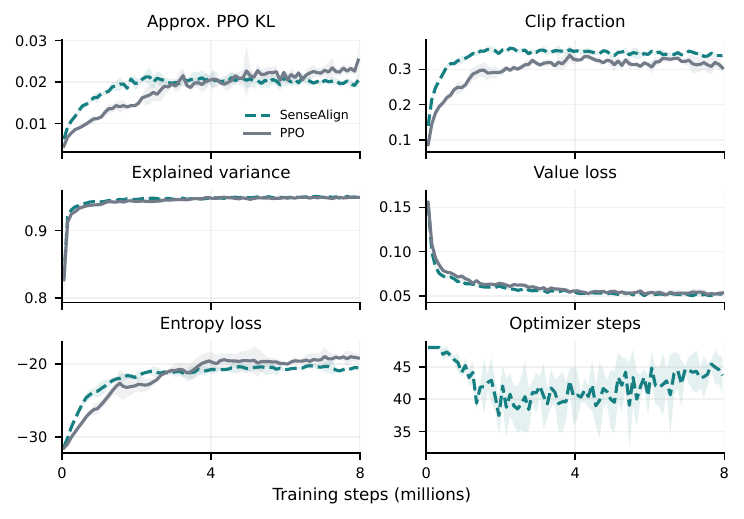}
\caption{PPO training metrics}
\end{subfigure}
\caption{Training logs for congestion-change monitoring: shaded bands show the sample SD among training seeds.}
\label{fig: training logs}
\end{figure}

\paragraph{Detailed Experiment Records}
Table~\ref{tab: congestion all seeds} contains the congestion-change monitoring reward of various configurations.
SenseAlign outperforms PPO consistently across all the training random seeds, and the overall performance remains robust for a wide range of loss coefficients.
Table~\ref{tab: mfd error all seeds} shows the per-seed ablation studies on MFD estimation, and Figure~\ref{fig: reward curves all seeds} shows the reward curve by random seeds.
In both cases, SenseAlign has a consistent advantage.
Figure~\ref{fig: training logs} contains the training logs of SenseAlign and PPO, which shows that the training pipeline is stable.

\begin{table}[!htb]
\centering
\caption{Congestion-change return by training seed for various configurations.
The overall result reports the mean $\pm$ sample SD.}
\label{tab: congestion all seeds}
\begin{tabular}{lrrrrrrr}
\toprule
Configuration & $c_{\mathrm{pred}}$ & $c_{\mathrm{dec}}$ & $c_{\mathrm{KD}}$
& Seed 10086 & Seed 10087 & Seed 10088
& Overall $\uparrow$ \\
\midrule
SenseAlign               & 0.0001 & 0.1 & 0.1 & 31.22 & 30.74 & 31.63 & 31.20 $\pm$ 0.45 \\
w/o $\Lpred$       & 0      & 0.1 & 0.1 & 30.62 & 30.43 & 31.28 & 30.77 $\pm$ 0.45 \\
w/o $\Ldec$        & 0.0001 & 0   & 0.1 & 30.39 & 30.42 & 31.09 & 30.63 $\pm$ 0.40 \\
w/o $\Lkd$         & 0.0001 & 0.1 & 0   & 29.60 & 28.69 & 30.10 & 29.46 $\pm$ 0.71 \\
PPO                       & 0      & 0   & 0   & 28.72 & 28.47 & 29.37 & 28.85 $\pm$ 0.46 \\
\midrule
$c_{\mathrm{pred}}=0.001$ & 0.001  & 0.1 & 0.1 & 31.28 & 30.83 & 31.37 & 31.16 $\pm$ 0.29 \\
$c_{\mathrm{pred}}=0.01$  & 0.01   & 0.1 & 0.1 & 30.82 & 30.49 & 31.26 & 30.86 $\pm$ 0.39 \\
$c_{\mathrm{pred}}=0.1$   & 0.1    & 0.1 & 0.1 & 30.61 & 30.61 & 31.73 & 30.98 $\pm$ 0.65 \\
$c_{\mathrm{dec}}=0.5$    & 0.0001 & 0.5 & 0.1 & 30.70 & 30.29 & 30.45 & 30.48 $\pm$ 0.21 \\
$c_{\mathrm{dec}}=1.0$    & 0.0001 & 1.0 & 0.1 & 30.34 & 30.95 & 32.09 & 31.13 $\pm$ 0.89 \\
$c_{\mathrm{KD}}=0.2$     & 0.0001 & 0.1 & 0.2 & 30.56 & 30.36 & 31.02 & 30.65 $\pm$ 0.34 \\
$c_{\mathrm{KD}}=1.0$     & 0.0001 & 0.1 & 1.0 & 29.91 & 30.22 & 31.00 & 30.38 $\pm$ 0.57 \\
$c_{\mathrm{KD}}=2.0$     & 0.0001 & 0.1 & 2.0 & 29.94 & 30.70 & 31.20 & 30.61 $\pm$ 0.63 \\
\midrule
Privileged teacher       & --     & --  & --  & 35.53 & 34.12 & 37.05 & 35.57 $\pm$ 1.47 \\
\bottomrule
\end{tabular}
\end{table}

\begin{table}[!htb]
\centering
\caption{MFD estimation error (\%) by training seed for various configurations. The overall result reports the mean $\pm$ sample SD.}
\label{tab: mfd error all seeds}
\begin{tabular}{lrrrrrrr}
\toprule
Configuration & $c_{\mathrm{pred}}$ & $c_{\mathrm{dec}}$ & $c_{\mathrm{KD}}$
& Seed 10086 & Seed 10087 & Seed 10088
& Overall $\downarrow$ \\
\midrule
SenseAlign & 0.0001 & 0.1 & 0.1 & 5.54 & 5.55 & 6.02 & 5.70 $\pm$ 0.27 \\
w/o $\Lpred$ & 0 & 0.1 & 0.1 & 5.62 & 5.77 & 6.08 & 5.82 $\pm$ 0.23 \\
w/o $\Ldec$ & 0.0001 & 0 & 0.1 & 5.59 & 5.59 & 6.13 & 5.77 $\pm$ 0.31 \\
w/o $\Lkd$ & 0.0001 & 0.1 & 0 & 5.46 & 6.51 & 6.56 & 6.18 $\pm$ 0.62 \\
PPO & 0 & 0 & 0 & 6.37 & 6.21 & 6.70 & 6.43 $\pm$ 0.25 \\
\midrule
Privileged teacher & -- & -- & -- & 5.43 & 4.80 & 5.02 & 5.08 $\pm$ 0.32 \\
\bottomrule
\end{tabular}
\end{table}

\section{Discussion and conclusion}
\label{sec: discussion and conclusion}
This paper presents SenseAlign, an RL-based framework that learns predictive representations for drone-based active sensing.
We address the problem in a partially observable large-scale urban road network, where the sensing utility is grounded in the complex spatio-temporal traffic dynamics. 
A privileged teacher is first trained to guide a student's actions directly and encourages the student's predictions to keep decision-relevant forecasts.
The student model then combines future traffic prediction, knowledge distillation, and decision alignment to jointly shape its latent space, which eventually enhances the actor's capability without any inference-time cost.

Experiments with the dense urban road network of central Barcelona have demonstrated improvements in active sensing using both congestion-change monitoring and MFD estimation as objectives.
With ten drones providing $\sim$5\% spatial coverage, SenseAlign improves congestion-monitoring by 8.1\% over PPO and reduces MFD estimation error from 6.43\% to 5.70\%.
Detailed results show that the performance gains persist across diverse traffic demands, and congestion-monitoring at most locations is improved due to better visit timing. 
Ablation studies confirmed that all three auxiliary objectives contribute to performance, and the policy learning remains robust under a reasonable hyperparameter range.

\section*{Reproducibility and AI assistance}
An AI assistant is used for improving visualization aesthetics and polishing language.
All assisted content has been checked and revised by the authors, who remain responsible for the final manuscript.
This work is built upon publicly available data and code, and will be made fully available once the paper is accepted for publication.

\section*{Acknowledgements}
This work is supported by the Swiss National Science Foundation (SNSF) under NCCR Automation, grant agreement number 51NF40\_180545.

\bibliography{references}
\bibliographystyle{elsarticle-harv}

\end{document}